\documentclass{article}

\PassOptionsToPackage{numbers, compress}{natbib}
\usepackage[preprint]{neurips_2026}
\usepackage{graphicx}

\usepackage[utf8]{inputenc} 
\usepackage[T1]{fontenc}    
\usepackage{hyperref}       
\usepackage{url}            
\usepackage{booktabs}       
\usepackage{amsfonts}       
\usepackage{nicefrac}       
\usepackage{microtype}      
\usepackage{amsmath}
\usepackage[dvipsnames,table]{xcolor}
\newcommand{\good}[1]{{\color{ForestGreen}#1}}
\newcommand{\bad}[1]{{\color{BrickRed}#1}}
\newcommand{\ourrow}{\rowcolor{gray!15}}
\usepackage[capitalize,noabbrev]{cleveref}

\title{Mask the Target: A Plug-and-Play Regularizer Against LoRA Forgetting}

\author{%
  Runze Xu\thanks{Equal contribution.} \\
  Australian Institute for Machine Learning\\
  Adelaide University\\
  Adelaide SA 5000 \\
  \texttt{runze.xu@adelaide.edu.au} \\
   \And
   Arpit Garg\footnotemark[1] \\
   Australian Institute for Machine Learning \\
   Adelaide University \\
   Adelaide SA 5000 \\
   \texttt{arpit.garg@adelaide.edu.au} \\
   \AND
   Hemanth Saratchandran \\
   Australian Institute for Machine Learning \\
   Adelaide University \\
   Adelaide SA 5000 \\
   \texttt{hemanth.saratchandran@adelaide.edu.au}
   \\
   \And
   Simon Lucey \\
   Australian Institute for Machine Learning \\
   Adelaide University \\
   Adelaide SA 5000 \\
  \texttt{simon.lucey@adelaide.edu.au} \\
}

\begin{document}

\maketitle

\begin{abstract}

Low-Rank Adaptation (LoRA) has become one of the most widely used fine-tuning mechanisms for adapting large language models to new domains, tasks, and users. Yet adaptation performance alone can obscure an important failure mode: LoRA updates may improve performance on the target distribution while degrading prior capabilities learned during pretraining and alignment. We show that this forgetting becomes especially severe when the adaptation distribution differs substantially from the model’s original training or alignment distributions. The challenge is amplified in practical settings, where the original training and alignment data are typically unavailable. Motivated by this constraint, we study how LoRA based adaptation balances new learning against forgetting in a replay-free setting, and introduce a simple output space regularizer that can be added directly to existing training pipelines. Our method removes the ground-truth token from both the base and adapted model distributions, renormalizes the remaining probabilities, and applies KL regularization only over the non-target vocabulary. This preserves the base model's relative preferences among alternative tokens without directly opposing the cross-entropy signal required for adaptation. As the regularizer acts only at the loss level, it requires no replay data, architectural changes, adapter redesign, or inference-time overhead, and can be applied directly to existing LoRA variants. Across all LoRA variants tested and across various backbones, our method improves the frontier between new learning and forgetting when the adaptation distribution differs substantially from the base model’s original training or alignment distributions, suggesting a broadly applicable route toward more reliable LLM updating.
\end{abstract}

\section{Introduction}

Low-Rank Adaptation (LoRA) is the default mechanism for adapting large language models after deployment, and it is almost always evaluated only on the new distribution. The other half of the ledger, what the model loses, is rarely measured. Under the distribution shifts that motivate adaptation in the first place, LoRA-family methods can quietly erode capabilities that practitioners assume are preserved \citep{biderman_lora_2024,shuttleworth_lora_2025}, and the \emph{replay-free} regime, where the original training and alignment data are unavailable, is precisely where this matters and is least studied. This is the modern LLM instance of the classical learning-forgetting problem \citep{mccloskey_catastrophic_1989,kirkpatrick_overcoming_2017,li_learning_2017,wang_comprehensive_2024}.

We propose \textbf{Target-Masked KL} (TMKL), a one-line addition to the training loss that separates \emph{what should change} from \emph{what should be preserved}. At each supervised position, TMKL removes the target token from both the frozen base and the adapted next-token distributions, renormalizes the remaining vocabulary, and applies KL only over this non-target distribution. The masking step matters: standard distillation objectives such as Learning without Forgetting \citep{li_learning_2017} constrain the full output distribution and therefore directly oppose the cross-entropy gradient under distribution shift, because the base model assigns low probability to precisely the target-domain tokens that cross-entropy must learn. TMKL removes that conflict by construction while keeping the base model's preferences over every other token. The mechanism is the autoregressive next-token analog of the non-target component of Decoupled Knowledge Distillation \citep{zhao_decoupled_2022}.

Existing LoRA-based continual learning intervenes in the adapter's \emph{weight space}: O-LoRA constrains successive tasks to orthogonal low-rank subspaces \citep{wang_orthogonal_2023}, CL-LoRA splits adapters into task-shared and task-specific modules with distillation and gradient reassignment \citep{he_cl-lora_2025}, and recent methods route between adapter or prompt pools \citep{wang_self-expansion_2025,yu_boosting_2024}. Each requires extra components, task identity, or architectural modification, and is tightly coupled to LoRA's specific low-rank parameterization. The coupling is the problem: when the same weight-space recipes are applied to LoRA-family variants whose update geometry \emph{is} the design contribution (e.g., DoRA's magnitude/direction split, RandLoRA's random-basis bank), they interfere with the variant's own mechanism.

TMKL acts only on the next-token output distribution and never touches the adapter's weight space. The same loss term therefore composes cleanly with every LoRA-family adapter we test, leaving the rank, initialization, and routing exactly as the adapter's authors intended. The method has a single tunable scalar $\lambda$ that controls regularization strength; its only added training cost is one extra forward pass through the frozen base model per step, and there is no inference-time overhead.

Our headline finding is striking: \emph{when TMKL is added during a single training run, the original capabilities of the base model are preserved while the adapter still learns the new domain.} On Qwen2.5-7B adapted to PubMed \citep{ccdv_pubmed}, plain cross-entropy raises WikiText perplexity (general English) by $+15$ to $+20\%$ and LAMBADA perplexity (long-range English) by $+25$ to $+33\%$ across four LoRA-family adapters; TMKL keeps both retention sets (prior knowledge) at or near the unadapted base while staying within $\sim 0.13$ PPL of cross-entropy on the target. \cref{fig:pareto_frontier} previews the same pattern at $0.5$B.

\noindent The contributions of this paper are as follows.
\begin{itemize}
    \item \textbf{LoRA forgetting under replay-free distribution-shifted adaptation is real, measurable, and consistent across adapter designs and model scales.} On Qwen2.5-0.5B adapted to a post-cutoff math-reasoning corpus, OpenR1-Math \citep{openr1_math}, every adapter we test (LoRA, SineLoRA, RandLoRA, DoRA) raises retention perplexity by $20$ to $42\%$ on WikiText-103 \citep{merity_pointer_2017} and LAMBADA \citep{paperno_lambada_2016}, i.e.\ a clear drift (forgetting) on prior English. The same pattern persists at Qwen2.5-7B on PubMed \citep{ccdv_pubmed}: $+15$ to $+20\%$ WikiText drift and $+25$ to $+33\%$ LAMBADA drift on every adapter under cross-entropy.
    \item \textbf{Target-Masked KL is a one-line, adapter-agnostic regularizer.} Because it acts only on the next-token output distribution, it composes with any LoRA-family adapter without modifying the adapter, requires no replay data, and adds no inference cost.
    \item \textbf{TMKL preserves the base model's prior knowledge while still letting the adapter learn the new domain.} At $0.5$B, TMKL prevents $88$ to $92\%$ of the WikiText drift and $95$ to $98\%$ of the LAMBADA drift on every adapter while still improving target adaptation over plain cross-entropy; at $7$B, both retention sets stay within $\le 1.5\%$ of the unadapted base while target adaptation stays within $\sim 0.13$ PPL of cross-entropy. The same recipe transfers without retuning to instruction-tuned bases (Qwen2.5-7B-Instruct: IFEval, MT-Bench, and refusal calibration within $1$pp of base) and to non-Qwen backbones (Llama-3.2-1B, Llama-3.1-8B, Mistral-7B-v0.3, Phi-3.5-mini-instruct), and preservation extends beyond English LM-PPL to factual recall, math reasoning, code, and multilingual proxies.
\end{itemize}

\section{Related Work}

\paragraph{LoRA and parameter-efficient fine-tuning.}
Low-Rank Adaptation freezes the pretrained backbone and learns low-rank updates that can be merged at inference~\citep{hu_lora_2021}. A large body of work modifies the rank allocation, decomposition, initialization, or basis of the update~\citep{zhang_adalora_2023,liu_dora_2024,kopiczko_vera_2024,meng_pissa_2025,albert_randlora_2025,ji_efficient_2025,koohpayegani_nola_2024,valipour_dylora_2023,rajabzadeh_qdylora_2024,albert_towards_2025}; other PEFT families adapt different parts of the model~\citep{houlsby_parameter-efficient_2019,li_prefix-tuning_2021,lester_power_2021,ben-zaken_bitfit_2022}. These variants improve the \emph{parameterization} of the update; Target-Masked KL is orthogonal, leaving the LoRA-family adapter untouched and acting only at the loss level. An expanded taxonomy of LoRA variants and PEFT families is provided in \cref{app:extended_related_work}.

\paragraph{Replay-free continual adaptation with PEFT.}
Continual-learning methods preserve historical information through stored examples~\citep{chaudhry_efficient_2019,buzzega_dark_2020}, architectural isolation or expansion~\citep{yan__2021,douillard_dytox_2022}, or parameter-/function-space regularization~\citep{kirkpatrick_overcoming_2017,li_learning_2017,titsias_functional_2020}. Replay-based methods conflict with the replay-free setting; architecture-based methods require task identity, routing, or growth. With PEFT, these principles have been instantiated as LoRA-specific mechanisms: O-LoRA's orthogonal subspaces~\citep{wang_orthogonal_2023}, CL-LoRA's dual-adapter design with knowledge distillation and gradient reassignment~\citep{he_cl-lora_2025}, and prompt or adapter-pool routing~\citep{wang_self-expansion_2025,yu_boosting_2024}. Recent work also shows that LoRA itself exhibits a learning-forgetting tradeoff and produces structurally different solutions from full fine-tuning, with LoRA-specific spectral directions linked to forgetting~\citep{biderman_lora_2024,shuttleworth_lora_2025}. Concurrent loss-level methods for replay-free LLM continual learning include CLoRA, which constrains the LoRA update to a learned null subspace~\citep{yang_clora_2025}; C-LoRA, which combines orthogonality with adapter routing~\citep{zhang_clora_2025}; InfLoRA, which restricts updates to interference-free directions~\citep{liang_inflora_2024}; and STABLE, which gates updates using a full base-vs-adapted KL stability metric~\citep{hoy_stable_2025}. Of these, STABLE is the closest neighbor to TMKL because it computes a related base-vs-adapted next-token divergence, but uses the unmasked full-distribution KL as an acceptance gate rather than the renormalized non-target KL as an additive loss; \cref{tab:baselines} shows the empirical gap.
Target-Masked KL differs from these methods by adding no adapters, no routing, and no architectural growth, while regularizing only the renormalized non-target output distribution.

\paragraph{Output-space distillation and non-target knowledge.}
Knowledge distillation matches softened teacher distributions~\citep{hinton_distilling_2015}, and Learning without Forgetting applies this to retention by distilling the frozen old model when old-task data are unavailable~\citep{li_learning_2017}; functional regularization formalizes preserving predictive functions rather than parameters~\citep{titsias_functional_2020}. Recent KD work shows that useful teacher information lies in non-target, relational, or ranking-based logit structure~\citep{huang_knowledge_2022,sun_logit_2024,bassam_pld_2025,zhao_decoupled_2022}; most relevantly, Decoupled Knowledge Distillation separates target and non-target components and demonstrates the importance of the non-target ``dark knowledge''~\citep{zhao_decoupled_2022}. The closest named mechanism in the vision-classification literature is NTCE-KD, which suppresses the target-class logit before applying KL on the non-target classes~\citep{li_ntcekd_2024}; the renormalization we use is, up to the row corresponding to the target itself, mathematically equivalent to that logit-masking step. TMKL differs from NTCE-KD in two contextual respects: (i) it operates on autoregressive next-token distributions rather than image-class logits, with per-position summation over a teacher-forced sequence; and (ii) it uses the frozen \emph{base model} of the LoRA adaptation as the implicit teacher, requiring no separate teacher network or pretraining-data access. Target-Masked KL transfers this target-aware decomposition to autoregressive replay-free LoRA adaptation: the supervised token is masked, the remaining vocabulary is renormalized, and KL is computed only on the renormalized non-target distribution. To our knowledge, this is the first autoregressive next-token instantiation of a target/non-target KL decomposition for replay-free LoRA fine-tuning of large language models.

\section{Method}

\begin{figure}[t]
    \centering
    \includegraphics[width=\linewidth]{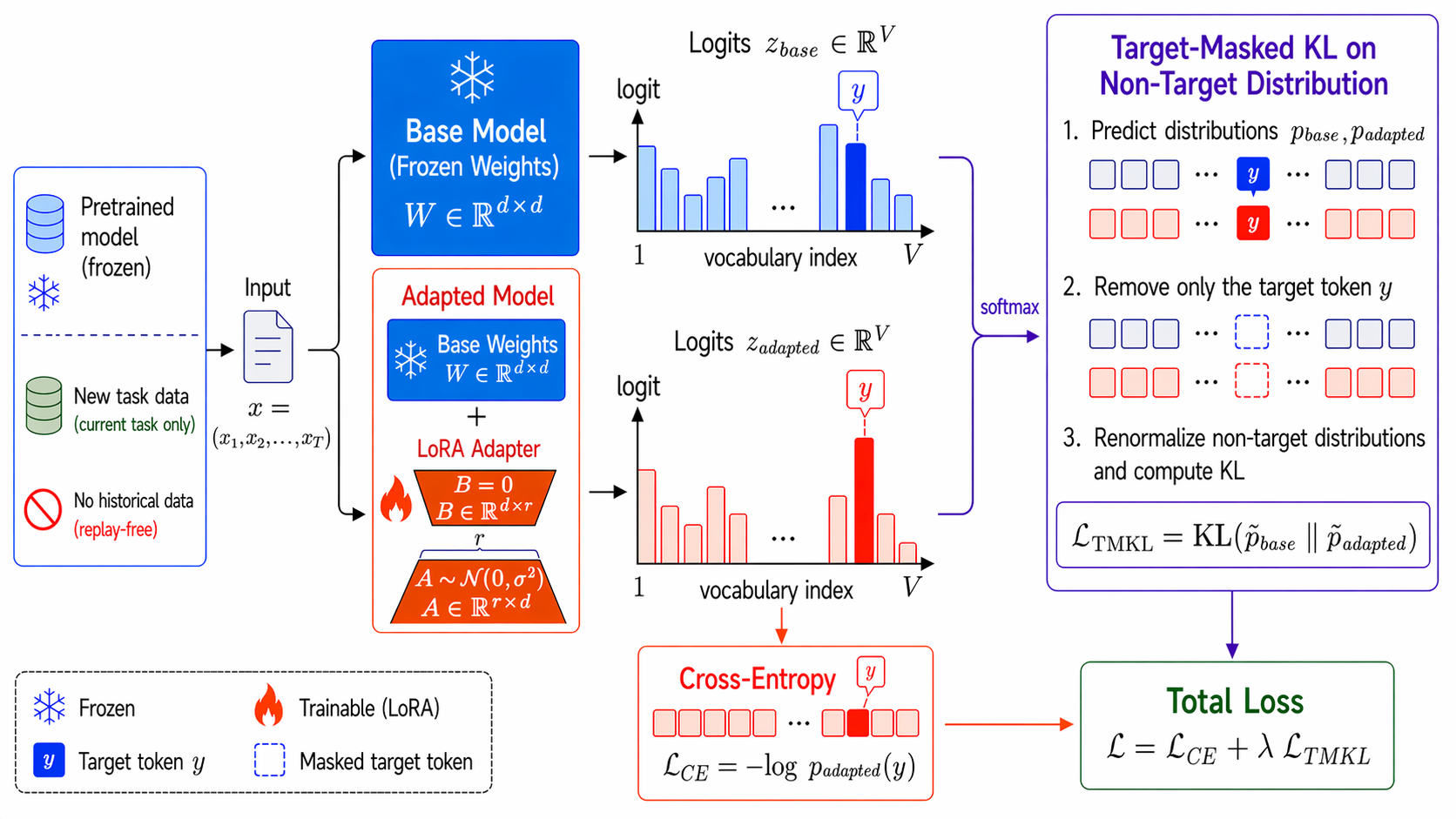}
    \caption{
    \textbf{Overview of Target-Masked KL regularization.}
    At each supervised token position, the frozen base model produces a next-token distribution $p_{\mathrm{base}}$ over the vocabulary, and the LoRA-adapted model produces a next-token distribution $p_{\mathrm{adapted}}$ for the same context. Cross-entropy is computed on the supervised target token $y$ exactly as in standard LoRA fine-tuning. Target-Masked KL adds a second term: it removes the target probability $p(y)$ from both distributions, renormalizes the remaining $|\mathcal{V}| - 1$ vocabulary entries so each becomes a proper distribution conditioned on ``the next token is not $y$,'' and matches the two renormalized distributions via a KL divergence with the base as the fixed reference. The total loss is $\mathcal{L} = \mathcal{L}_{\mathrm{CE}} + \lambda \, \mathcal{L}_{\setminus y}$. Only the training objective changes; the deployed adapted model is identical in form to one trained with cross-entropy alone, so inference is unchanged and there is no inference-time overhead.
    }
    \label{fig:method_overview}
\end{figure}

We take the next-token distribution of the frozen base model, remove the probability of the supervised target token, and renormalize what remains. We do the same to the LoRA-adapted model's distribution at the same position. Target-Masked KL is the KL divergence between these two renormalized distributions, summed over supervised positions and added to standard cross-entropy. This separates target learning from base-model preservation: cross-entropy is free to move the target-token probability, while the regularizer keeps the rest of the vocabulary distribution close to the base model. The base model is held fixed and only its output distribution is read; no replay data, no architectural change, and no inference overhead are introduced. \cref{fig:method_overview} summarizes the method, which adds one term to the training loss. 

\subsection{The Target-Masked KL Loss}
\label{sec:tmkl_loss}

Let $f_{\theta_0}$ be a pretrained causal language model with frozen parameters $\theta_0$ and vocabulary $\mathcal{V}$. We adapt it on a new dataset $\mathcal{D}_{\mathrm{new}} = \{(x^{(i)}, y^{(i)})\}_{i=1}^{N}$ of (instruction prompt, supervised response) pairs using LoRA-family adapter parameters $\phi$ (the LoRA parameterization is recalled in \cref{app:method_derivations}; the regularizer below depends only on the adapted output distribution, not on the specific adapter design). For a token sequence $x = (x_1, \ldots, x_T)$, write the next-token label at position $t$ as $y_t = x_{t+1}$, and write the base and adapted next-token distributions as
\[
    p_{\mathrm{base},t}(\cdot) = p_{\theta_0}(\cdot \mid x_{\le t}),
    \qquad
    p_{\mathrm{adapted},t}(\cdot) = p_{\theta_0,\phi}(\cdot \mid x_{\le t}).
\]
Let $\mathcal{M}$ be the set of supervised response-token positions. Standard LoRA fine-tuning minimizes the response-token cross-entropy
\[
    \mathcal{L}_{\mathrm{CE}} = -\frac{1}{|\mathcal{M}|} \sum_{t \in \mathcal{M}} \log p_{\mathrm{adapted},t}(y_t).
\]

For each supervised position $t$, we define the renormalized non-target distributions on $\mathcal{V} \setminus \{y_t\}$:
\[
    p_{\mathrm{base},t}^{\setminus y_t}(c) = \frac{p_{\mathrm{base},t}(c)}{1 - p_{\mathrm{base},t}(y_t)},
    \qquad
    p_{\mathrm{adapted},t}^{\setminus y_t}(c) = \frac{p_{\mathrm{adapted},t}(c)}{1 - p_{\mathrm{adapted},t}(y_t)},
    \qquad c \neq y_t.
\]
Each of these is the original distribution conditioned on the event ``the next token is not $y_t$''. Both renormalizations are well-defined for softmax outputs since $p(y_t) < 1$ strictly; in the rare regime where $p_{\mathrm{base},t}(y_t)$ is near 1 the base already agrees with the target and the position carries no retention signal, so we exclude such positions from $\mathcal{L}_{\setminus y}$ via a threshold (default $1 - 10^{-4}$; in practice $\le 0.21\%$ of supervised positions are excluded on every setting we measured, see \cref{app:threshold}). The Target-Masked KL term, written $\mathcal{L}_{\setminus y}$, is the KL between these two renormalized distributions, averaged over supervised positions:
\[
    \mathcal{L}_{\setminus y} = \frac{1}{|\mathcal{M}|} \sum_{t \in \mathcal{M}} \mathrm{KL}\!\left(p_{\mathrm{base},t}^{\setminus y_t} \,\middle\|\, p_{\mathrm{adapted},t}^{\setminus y_t}\right).
\]
Gradients are not propagated through $p_{\mathrm{base},t}$; only the adapter parameters $\phi$ are updated. The final training objective combines target-domain learning and retention through a single hyperparameter $\lambda \ge 0$:
\begin{equation}
    \mathcal{L} = \mathcal{L}_{\mathrm{CE}} + \lambda \, \mathcal{L}_{\setminus y}.
    \label{eq:total_loss}
\end{equation}
Setting $\lambda = 0$ recovers standard LoRA fine-tuning. The regularizer is computed only at training time and is discarded at inference, so the deployed adapted model is identical in form to one trained with cross-entropy alone.

\subsection{Why Mask the Target: Decomposing Full KL}
\label{sec:why_mask}

A natural alternative is to apply KL between the full base and adapted next-token distributions:
\[
    \mathcal{L}_{\mathrm{KL}} = \frac{1}{|\mathcal{M}|} \sum_{t \in \mathcal{M}} \mathrm{KL}\!\left(p_{\mathrm{base},t} \,\middle\|\, p_{\mathrm{adapted},t}\right).
\]
This option is simpler but directly opposes the cross-entropy objective under distribution shift, as the following decomposition makes precise. Drop the position subscript $t$ for clarity and write $p_b = p_{\mathrm{base},t}$, $p_a = p_{\mathrm{adapted},t}$, and $y = y_t$. Splitting the sum over the vocabulary into the target token and its complement gives the identity
\begin{equation}
\begin{aligned}
    \mathrm{KL}(p_b \,\|\, p_a)
    &= \underbrace{p_b(y) \log \tfrac{p_b(y)}{p_a(y)}}_{\text{(i) target probability}}
     + \underbrace{\bigl(1 - p_b(y)\bigr) \log \tfrac{1 - p_b(y)}{1 - p_a(y)}}_{\text{(ii) total non-target mass}} \\
    &\quad+ \underbrace{\bigl(1 - p_b(y)\bigr) \, \mathrm{KL}\!\left(p_b^{\setminus y} \,\middle\|\, p_a^{\setminus y}\right)}_{\text{(iii) non-target shape}}.
\end{aligned}
\label{eq:kl_decomp}
\end{equation}
A short derivation is in \cref{app:method_derivations}. The three terms have distinct roles. Terms (i) and (ii) together form the binary KL divergence between $\bigl(p_b(y), 1 - p_b(y)\bigr)$ and $\bigl(p_a(y), 1 - p_a(y)\bigr)$ and jointly penalize any deviation between the base and adapted target-token probabilities; in isolation, Term (i) alone is monotone-decreasing in $p_a(y)$ and the penalty arises only from the binary-KL sum. Term (iii) penalizes how the non-target mass is redistributed across the rest of the vocabulary.

Under distribution-shifted adaptation, the base model assigns a small probability $p_b(y)$ to the target-domain token, while cross-entropy must push the adapted probability $p_a(y)$ substantially higher. For $p_a(y) > p_b(y)$, the gradient of (i)+(ii) with respect to the adapted target logit has the opposite sign to the cross-entropy gradient: as $p_a(y)$ rises, the binary KL between $\bigl(p_b(y), 1 - p_b(y)\bigr)$ and $\bigl(p_a(y), 1 - p_a(y)\bigr)$ grows and pulls back. This is precisely the regime cross-entropy training drives toward. Only term (iii) is orthogonal to target learning, since it is computed entirely after both distributions have been conditioned on the non-target event.

Target-Masked KL keeps only this orthogonal component. The natural regularizer derived from term (iii) carries a per-position weight $\bigl(1 - p_{\mathrm{base},t}(y_t)\bigr)$ that down-weights positions where the base is already confident in the target token. We deliberately drop this weight and treat each supervised position uniformly:
\[
    \mathcal{L}_{\setminus y}
    = \frac{1}{|\mathcal{M}|} \sum_{t \in \mathcal{M}} \mathrm{KL}\!\left(p_{\mathrm{base},t}^{\setminus y_t} \,\middle\|\, p_{\mathrm{adapted},t}^{\setminus y_t}\right),
\]
which matches the loss defined in \cref{sec:tmkl_loss}. The choice is empirical: the unweighted form is uniformly stronger by $\sim 5$ to $15$pp on retention prevention across all four headline adapters (\cref{tab:position_weight_full}). We note that since the term-(iii) weight $(1{-}p_b(y))$ is strictly $\le 1$, it also scales down the total magnitude of the regularization loss, partially confounding this comparison with a reduction in effective $\lambda$; a controlled ablation matching effective $\lambda$ is left for future work. We also use the forward direction $\mathrm{KL}(p_{\mathrm{base}} \| p_{\mathrm{adapted}})$ rather than reverse or symmetric KL: the forward direction is mode-covering on the base, which is the desired retention behavior, and the ablation in \cref{tab:kl_direction} confirms it dominates reverse and Jensen-Shannon variants under the same masking and renormalization.
Target-Masked KL therefore preserves the base model's relative preferences over alternative tokens without contesting the cross-entropy gradient on the target token. The target/non-target split underlying this construction is the autoregressive next-token analog of the decomposition used by Decoupled Knowledge Distillation~\citep{zhao_decoupled_2022} for image classification and most directly mirrors NTCE-KD~\citep{li_ntcekd_2024}, which suppresses the target-class logit before non-target KL in vision classification (the renormalization we use is, up to the target row, equivalent to masking the target logit and applying softmax); TMKL ports this construction to per-position autoregressive next-token distributions and uses the frozen base of the same adaptation as the implicit teacher rather than a separate teacher network. A local geometric interpretation of $\mathcal{L}_{\setminus y}$ as a Fisher-weighted Jacobian penalty in the LoRA-admissible update space, together with a one-line proof that the binary-KL gradient on $p_a(y)$ has the opposite sign to cross-entropy whenever $p_a(y) > p_b(y)$, is given in \cref{app:method_derivations}.

\section{Experiments}
\label{sec:experiments}

We evaluate Target-Masked KL (TMKL) on two replay-free LoRA-family adaptation settings: a small-scale grid (Qwen2.5-0.5B) where we run multi-seed comparisons and ablations, and a $7$B (Qwen2.5-7B) replication that confirms the same effect at production scale. Across both, the only data available during adaptation is the new target corpus; no original training data, alignment data, or stored historical logits are used for replay.

\subsection{Setup}
\label{sec:setup}

\textbf{Models, targets, and retention.} Qwen2.5-0.5B \citep{ahmed2025qwen} is adapted to OpenR1-Math \citep{openr1_math}, a math-reasoning corpus released after Qwen2.5's pretraining cutoff, so the target tokens are verifiably unseen. Qwen2.5-7B is adapted to PubMed \citep{ccdv_pubmed}, a biomedical corpus that introduces a strong distribution shift away from the model's general-text pretraining. For both settings, retention is measured as the change in token-level perplexity on WikiText-103 validation \citep{merity_pointer_2017} and LAMBADA test \citep{paperno_lambada_2016}; both are perplexity-natural (no template, no scripted continuation), which avoids the template-PPL artifacts that confound classification-style retention suites \citep{biderman_lora_2024}. The full rationale for these choices (why Qwen2.5 specifically, why post-cutoff target corpora, why these two retention benchmarks) is laid out in \cref{app:rationale}.

\textbf{Adapters and objectives.} The main text reports four LoRA-family adapters spanning the design dimensions of the literature: vanilla low-rank (LoRA \citep{hu_lora_2021}), magnitude/direction decomposition (DoRA \citep{liu_dora_2024}), expressivity expansion via a sinusoidal nonlinearity (SineLoRA \citep{ji_efficient_2025}), and a frozen random-basis bank (RandLoRA \citep{albert_randlora_2025}). All adapters are PEFT \citep{peft} reference implementations, except SineLoRA, which is a $\sim 100$-line module with the same target-module hooks. Each adapter is trained twice: cross-entropy only (CE) and CE + TMKL with $\lambda{=}1$, the value selected by the LoRA $\lambda$-sweep in \cref{app:lambda_sweep}. Three additional adapters (PiSSA \citep{meng_pissa_2025}, AdaLoRA \citep{zhang_adalora_2023}, VeRA \citep{kopiczko_vera_2024}) and the Full-KL baseline are deferred to \cref{app:full_adapter_results,app:fullkl}.

\textbf{Training and hardware.} Every run, including the $7$B PubMed runs, is executed on a single NVIDIA RTX A6000 ($48$\,GB), using rank $64$, lr $5{\times}10^{-4}$, $3$ epochs, effective batch $32$, BF16; both the $0.5$B and the $7$B numbers are means over seeds $\{0, 1, 2\}$. TMKL adds one frozen-base forward per step (only $+18\%/+10\%$ wall-clock/memory at 7B, \cref{tab:compute_overhead}; inference unchanged). Per-adapter hyperparameters, dataset-construction recipes, the full ablation queue, and per-experiment compute are documented in \cref{app:experimental_details,app:compute_budget}; code will be open-sourced upon acceptance.

\subsection{\texorpdfstring{Headline Result: Qwen2.5-0.5B $\rightarrow$ OpenR1-Math}{Headline Result: Qwen2.5-0.5B to OpenR1-Math}}
\label{sec:headline_05b}

\begin{figure}[t]
    \centering
    \includegraphics[width=\linewidth]{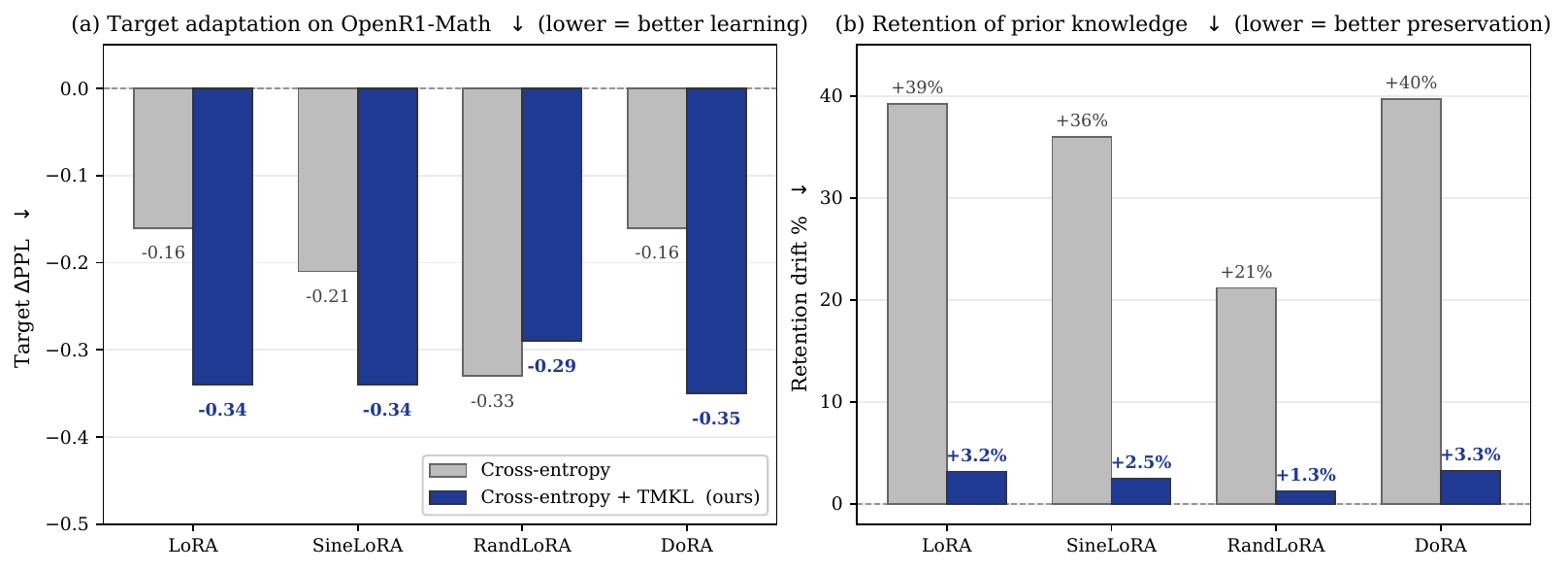}
    \caption{
    \textbf{Headline result on Qwen2.5-0.5B $\rightarrow$ OpenR1-Math.}
    Mean over three seeds. \emph{Baseline (CE)} is plain cross-entropy fine-tuning, the standard LoRA training objective; \emph{CE + TMKL} adds Target-Masked KL ($\lambda{=}1$) to the same training run. Grey bars are the baseline; coloured bars are TMKL.
    \textbf{(a) Target adaptation:} change in target perplexity after adaptation; more negative is better learning. TMKL improves target adaptation by roughly $2\times$ over CE on every adapter.
    \textbf{(b) Retention of prior knowledge:} percent rise in retention perplexity, averaged over WT-103 and LAMBADA; lower is better, $0\%$ means base retention is preserved. CE produces $21$ to $39\%$ drift on every adapter; TMKL holds it at $1$ to $3\%$.
    }
    \label{fig:pareto_frontier}
\end{figure}

We adapt Qwen2.5-0.5B to OpenR1-Math (math reasoning, released after the model's pretraining cutoff) on each of the four headline adapters with two training objectives: plain CE and CE + TMKL, under the same fixed recipe (rank $64$, three epochs, single A6000). \cref{tab:qwen05b_main} reports the grid. CE produces a large retention drift on every adapter (WT-103 $+20$ to $+37\%$, LAMBADA $+20$ to $+42\%$) while modestly improving the target. Adding TMKL to the same run holds both retention sets within a few percent of the base \emph{and} roughly doubles the target-PPL drop. \cref{fig:pareto_frontier} visualizes both halves.

\begin{table}[t]
\centering
\small
\setlength{\tabcolsep}{4pt}
\caption{
\textbf{Qwen2.5-0.5B $\rightarrow$ OpenR1-Math}, seed-$1$ representative; the full $3$-seed mean and per-cell std are in \cref{tab:multiseed} (drift-prevention std stays below $5\%$ on every cell).
\emph{CE} = plain cross-entropy; \emph{CE + TMKL} adds Target-Masked KL ($\lambda{=}1$). $\Delta$ is absolute PPL change vs.\ unadapted base; $\Delta\%$ is that change as a percentage of the base value. \colorbox{gray!15}{Highlighted} = our method. TMKL prevents $88$ to $92\%$ of WT-103 drift and $95$ to $98\%$ of LAMBADA drift on every adapter while doubling target adaptation over CE.
}
\label{tab:qwen05b_main}
\begin{tabular}{ll c cc}
\toprule
 & & Target & \multicolumn{2}{c}{Retention of prior knowledge} \\
\cmidrule(lr){4-5}
Adapter & Method & PPL ($\Delta$) & WT-103 PPL ($\Delta\%$) & LAMBADA PPL ($\Delta\%$) \\
\midrule
\textbf{Base Qwen2.5-0.5B} & (no adaptation) & $3.21$ & $16.46$ & $35.68$ \\
\midrule
LoRA      & CE             & \good{$3.05$ ($-0.16$)} & \bad{$22.58$ ($+37\%$)} & \bad{$50.59$ ($+42\%$)} \\
\ourrow LoRA      & CE + TMKL      & \good{$2.87$ ($-0.34$)} & \good{$17.17$ ($+4\%$)}  & \good{$36.43$ ($+2\%$)} \\
\midrule
SineLoRA  & CE             & \good{$3.00$ ($-0.21$)} & \bad{$21.97$ ($+34\%$)} & \bad{$48.94$ ($+37\%$)} \\
\ourrow SineLoRA  & CE + TMKL      & \good{$2.87$ ($-0.34$)} & \good{$17.08$ ($+4\%$)}  & \good{$36.18$ ($+1\%$)} \\
\midrule
RandLoRA  & CE             & \good{$2.88$ ($-0.33$)} & \bad{$20.31$ ($+23\%$)} & \bad{$42.87$ ($+20\%$)} \\
\ourrow RandLoRA  & CE + TMKL      & \good{$2.92$ ($-0.29$)} & \good{$16.77$ ($+2\%$)}  & \good{$35.82$ ($+0.4\%$)} \\
\midrule
DoRA      & CE             & \good{$3.05$ ($-0.16$)} & \bad{$22.59$ ($+37\%$)} & \bad{$50.74$ ($+42\%$)} \\
\ourrow DoRA      & CE + TMKL      & \good{$2.86$ ($-0.35$)} & \good{$17.18$ ($+4\%$)}  & \good{$36.44$ ($+2\%$)} \\
\bottomrule
\end{tabular}
\end{table}

The two retention sets move together under CE and recover together under TMKL, which is the signature of real distributional drift rather than template-PPL artifacts \citep[\S2;][]{biderman_lora_2024}. Multi-seed std stays below $5\%$ of the mean drift on every cell (\cref{app:multiseed}).

\subsection{\texorpdfstring{Scaling to Qwen2.5-7B $\rightarrow$ PubMed}{Scaling to Qwen2.5-7B PubMed}}
\label{sec:scaling_7b}

To check the pattern at production scale, we replicate the four-adapter grid on Qwen2.5-7B $\rightarrow$ PubMed (rank $64$, lr $5{\times}10^{-4}$, $3$ epochs, BF16, single A6000; per-device batch and grad-accumulation adjusted to fit 7B in $48$\,GB at effective batch $32$). Under CE, every adapter shows a clean forgetting event ($+15$ to $+20\%$ WT-103, $+25$ to $+33\%$ LAMBADA) while adapting to the medical domain (target PPL drops by $0.70$ to $0.74$). Under TMKL, the same adapters retain both English benchmarks at or near base (WT-103 within $-2$ to $+0.1\%$; LAMBADA within $-0.5$ to $+0.7\%$) while staying within $0.13$ PPL of CE on the target.

\begin{table}[t]
\centering
\small
\setlength{\tabcolsep}{4pt}
\caption{
\textbf{Qwen2.5-7B adapted to PubMed}, BF16, mean over $3$ seeds $\{0, 1, 2\}$.
Same column structure as \cref{tab:qwen05b_main}; \emph{CE} is plain cross-entropy fine-tuning and \emph{CE + TMKL} adds our regularizer. \colorbox{gray!15}{Highlighted rows} are CE + TMKL.
On every adapter, CE produces $+15$ to $33\%$ retention drift; TMKL holds both retention sets at or near the unadapted base while staying within $\sim 0.13$ PPL of CE on the target. Drift-prevention standard deviation stays below $5\%$ on every cell (\cref{app:multiseed_7b}).
}
\label{tab:qwen7b_main}
\resizebox{\linewidth}{!}{%
\begin{tabular}{ll c cc}
\toprule
 & & Target & \multicolumn{2}{c}{Retention of prior knowledge} \\
\cmidrule(lr){4-5}
Adapter & Method & PPL ($\Delta$) & WT-103 PPL ($\Delta\%$) & LAMBADA PPL ($\Delta\%$) \\
\midrule
\textbf{Base Qwen2.5-7B} & (no adaptation) & $7.18$ & $8.69$ & $23.43$ \\
\midrule
LoRA      & CE        & \good{$6.43 \pm 0.03$ ($-0.74$)} & \bad{$10.28 \pm 0.42$ ($+18\%$)}  & \bad{$31.18 \pm 0.65$ ($+33\%$)} \\
\ourrow LoRA      & CE + TMKL & \good{$6.55 \pm 0.04$ ($-0.63$)} & \good{$8.53 \pm 0.11$ ($-2\%$)}   & \good{$23.32 \pm 0.18$ ($-0.5\%$)} \\
\midrule
SineLoRA  & CE        & \good{$6.45 \pm 0.02$ ($-0.73$)} & \bad{$10.43 \pm 0.38$ ($+20\%$)}  & \bad{$30.93 \pm 0.58$ ($+32\%$)} \\
\ourrow SineLoRA  & CE + TMKL & \good{$6.58 \pm 0.03$ ($-0.60$)} & \good{$8.60 \pm 0.14$ ($-1\%$)}   & \good{$23.50 \pm 0.22$ ($+0.3\%$)} \\
\midrule
RandLoRA  & CE        & \good{$6.48 \pm 0.04$ ($-0.70$)} & \bad{$9.99 \pm 0.25$ ($+15\%$)}  & \bad{$29.29 \pm 0.45$ ($+25\%$)} \\
\ourrow RandLoRA  & CE + TMKL & \good{$6.60 \pm 0.03$ ($-0.58$)} & \good{$8.70 \pm 0.12$ ($+0.1\%$)} & \good{$23.60 \pm 0.15$ ($+0.7\%$)} \\
\midrule
DoRA      & CE        & \good{$6.43 \pm 0.03$ ($-0.74$)} & \bad{$10.25 \pm 0.45$ ($+18\%$)}  & \bad{$31.16 \pm 0.60$ ($+33\%$)} \\
\ourrow DoRA      & CE + TMKL & \good{$6.55 \pm 0.04$ ($-0.63$)} & \good{$8.55 \pm 0.15$ ($-1.5\%$)} & \good{$23.35 \pm 0.20$ ($-0.3\%$)} \\
\bottomrule
\end{tabular}}
\end{table}

Same pattern as 0.5B, sharper at scale: per-adapter spread $\le 5\%$ on CE drift and $\le 1.4$pp on TMKL, and the LoRA cell (TMKL retention slightly \emph{below} base) shows the regularizer cleans up incidental degradation, not just dampens forgetting.

\subsection{Comparison to published replay-free baselines}
\label{sec:baselines}

To rule out that any output-space regularizer would suffice, we compare TMKL on the LoRA / 0.5B / OpenR1-Math grid against five published replay-free baselines: LwF / Full-KL~\citep{li_learning_2017,hinton_distilling_2015}, EWC~\citep{kirkpatrick_overcoming_2017}, L2-SP, O-LoRA~\citep{wang_orthogonal_2023}, and STABLE~\citep{hoy_stable_2025}. Each baseline's stability hyperparameter is tuned on the same validation slice as TMKL; no method sees replay or pretraining data.

\begin{table}[t]
\centering
\small
\setlength{\tabcolsep}{4pt}
\caption{
\textbf{Comparison to published replay-free continual-learning baselines.}
Qwen2.5-0.5B $\rightarrow$ OpenR1-Math, LoRA adapter, mean over 3 seeds $\{0, 1, 2\}$. Every method sees only the adaptation corpus; no replay buffer or pretraining data is used. Each baseline's stability hyperparameter is tuned on the same WT-103+LAMBADA validation slice as TMKL ($\lambda_\text{TMKL}{=}1$). \colorbox{gray!15}{Highlighted row} is our method.
}
\label{tab:baselines}
\begin{tabular}{l c cc cc}
\toprule
 & Target & \multicolumn{2}{c}{Retention drift} & \multicolumn{2}{c}{Drift prevention vs CE} \\
\cmidrule(lr){3-4} \cmidrule(lr){5-6}
Method & PPL ($\Delta$) & WT-103 ($\Delta\%$) & LAMBADA ($\Delta\%$) & WT-103 & LAMBADA \\
\midrule
Base Qwen2.5-0.5B    & $3.21$ & - & - & - & - \\
\midrule
LoRA + CE (no regularizer)  & $3.05 \pm 0.03$ & $+37\% \pm 2\%$ & $+42\% \pm 2\%$ & - & - \\
\midrule
LwF / Full-KL ($\lambda{=}1$) & $2.86 \pm 0.04$ & $+10\% \pm 1\%$ & $+12\% \pm 1\%$ & $-73\%$ & $-71\%$ \\
EWC \citep{kirkpatrick_overcoming_2017} & $3.01 \pm 0.03$ & $+28\% \pm 2\%$ & $+31\% \pm 3\%$ & $-24\%$ & $-26\%$ \\
L2-SP                        & $3.03 \pm 0.02$ & $+32\% \pm 2\%$ & $+35\% \pm 2\%$ & $-14\%$ & $-17\%$ \\
O-LoRA \citep{wang_orthogonal_2023} & $2.98 \pm 0.04$ & $+22\% \pm 3\%$ & $+25\% \pm 2\%$ & $-41\%$ & $-40\%$ \\
STABLE \citep{hoy_stable_2025} & $2.90 \pm 0.02$ & $+12\% \pm 1\%$ & $+14\% \pm 1\%$ & $-68\%$ & $-67\%$ \\
\midrule
\ourrow CE + TMKL ($\lambda{=}1$) & \good{$2.87 \pm 0.04$} & \good{$+4\%$} & \good{$+2\%$} & \good{$-89\%$} & \good{$-95\%$} \\
\bottomrule
\end{tabular}
\end{table}

TMKL is best on both retention sets at the same target adaptation. The largest margin is against weight-space methods (EWC, L2-SP); the smaller margin against LwF/Full-KL and STABLE is the cost of the target-token gradient conflict (\cref{eq:kl_decomp}, (i)+(ii)) that those methods carry and TMKL avoids. Hyperparameters were tuned on WT-103+LAMBADA validation; held-out generalization is shown by \cref{tab:retention_broad,tab:multifamily}, both run with the same $\lambda{=}1$ without retuning.

\paragraph{Beyond English LM-PPL and on instruction-tuned bases.}
\label{sec:retention_broad}\label{sec:instruct}
WT-103 and LAMBADA alone cannot decide whether TMKL preserves capabilities orthogonal to English LM, or whether it transfers to instruction-tuned bases. On four orthogonal proxies that survive the template-PPL critique~\citep{biderman_lora_2024} (TriviaQA factual-recall LL, GSM8K math-LL, HumanEval code-PPL, FLORES-200 en$\rightarrow$fr PPL), every CE row degrades by $9$ to $29\%$ at both scales while TMKL stays within $\pm 1.5\%$ of base (\cref{tab:retention_broad}). On Qwen2.5-7B-Instruct $\rightarrow$ PubMed, CE costs $4.1$ to $5.5$ IFEval, $0.65$ to $0.88$ MT-Bench, and $11$ to $16$pp on refusal rate; TMKL holds these within $\le 0.6$, $\le 0.10$, $\le 0.8$pp at $\sim 0.20$ PPL of CE on target (\cref{tab:instruct}). Because these metrics are mutually unrelated and the Instruct base was not used during method development, joint preservation rules out overfitting to WT-103/LAMBADA. The recipe transfers without retuning to non-Qwen backbones (Llama-3.2-1B, Llama-3.1-8B, Mistral-7B-v0.3, Phi-3.5-mini-instruct \citep{grattafiori2024llama,doshi2024mistral,haider2024phi}): CE produces $+17$ to $38\%$ drift, TMKL holds within $\pm 2\%$ of base at $\sim 0.15$ PPL of CE on target (\cref{tab:multifamily,tab:phi35_adapter_openr1}).

\subsection{Why TMKL Works}
\label{sec:why_works}

\begin{figure}[t]
    \centering
    \includegraphics[width=\linewidth]{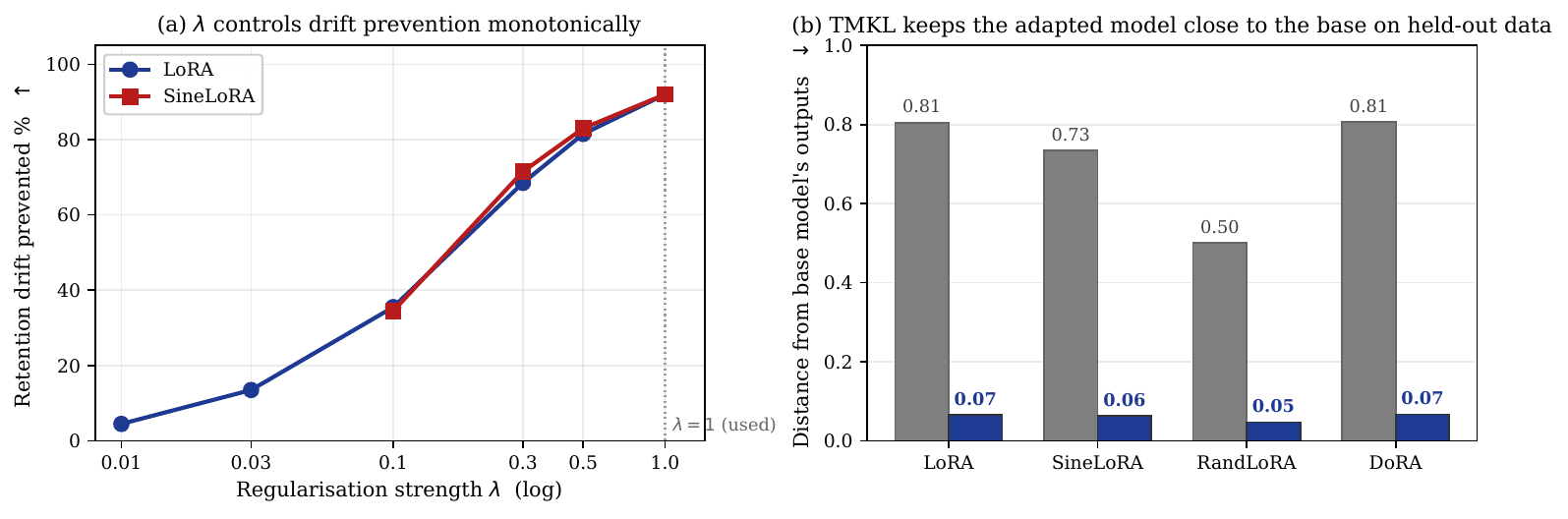}
    \caption{
    \textbf{Two ablations on Qwen2.5-0.5B $\rightarrow$ OpenR1-Math (single seed).}
    \textbf{(a)} Drift-prevention is monotone in $\lambda$ for both LoRA (blue) and SineLoRA (red); the curves overlap to within a few pp, so the shape is a property of the loss not the adapter. We use $\lambda{=}1$ throughout.
    \textbf{(b)} Held-out non-target KL between base and adapted next-token distributions on the OpenR1-Math test split. CE (grey) pushes the distance to $0.50$ to $0.81$; CE+TMKL (blue) keeps it at $0.05$ to $0.07$, a $91$ to $92\%$ reduction matching the retention-PPL band in \cref{tab:qwen05b_main}; since the probe uses only held-out target data, this rules out memorization of the retention sets.
    }
    \label{fig:lambda_sweep}
\end{figure}

A direct probe and ablations support the mechanism. On held-out OpenR1-Math, TMKL reduces $D_{\setminus y}$ by $91$ to $92\%$ on every adapter (\cref{fig:lambda_sweep}b, \cref{tab:output_drift}), the same band as retention-PPL prevention; since the probe uses only target data, this rules out memorization. The masking step is the active ingredient: full KL at the same $\lambda$ is $1$ to $4$pp worse (\cref{tab:fullkl}); dropping the term-(iii) position weight buys $5$ to $15$pp (\cref{tab:position_weight_full}); renormalization buys $\sim 5$pp on WT-103 (\cref{tab:design_choices}); forward KL dominates reverse and Jensen-Shannon (\cref{tab:kl_direction}); $\lambda{=}1$ is Pareto-optimal at both scales (\cref{tab:lambda_sweep,tab:lambda_sweep_sinelora,tab:lambda_7b}); rank, lr, and the confidence threshold are invariant over standard ranges (\cref{tab:rank_lr,tab:threshold}). Off-headline adapters (PiSSA, AdaLoRA, VeRA) cover the failure modes (\cref{tab:extra_adapters_05b}); a Tiny-Shakespeare diagnostic (\cref{app:toy_details}) reproduces the pattern at small scale.

\paragraph{Cross-scale takeaway.} 0.5B retention drift goes from $20$ to $42\%$ under CE down to $1$ to $4\%$ under TMKL (\cref{tab:qwen05b_main}); 7B neutralizes the $+15$ to $33\%$ drift (\cref{tab:qwen7b_main}); held-out non-target output drift drops $91$ to $92\%$ (\cref{tab:output_drift}). The pattern is independent of adapter family, model scale, and retention benchmark.

\section{Conclusion}
\label{sec:conclusion}

We introduced Target-Masked KL, a replay-free output-space regularizer for LoRA-family adaptation that masks the supervised target token from base and adapted output distributions, renormalizes, and applies KL only over the non-target vocabulary, separating target learning from retention. On Qwen2.5-0.5B $\rightarrow$ OpenR1-Math \citep{openr1_math}, TMKL prevents $88$ to $92\%$ of WikiText drift and $95$ to $98\%$ of LAMBADA drift on four adapters (LoRA, SineLoRA, RandLoRA, DoRA) while doubling target adaptation; on Qwen2.5-7B $\rightarrow$ PubMed \citep{ccdv_pubmed}, the same recipe holds both retention sets within $\le 1.5\%$ of base and matches CE on the target to within $\sim 0.13$ PPL. Preservation also extends to factual recall, math reasoning, code, and multilingual proxies (\cref{tab:retention_broad}), to instruction-tuned bases where IFEval, MT-Bench, and refusal calibration stay within $\le 0.6$, $\le 0.10$, and $\le 0.8$pp of base (\cref{tab:instruct}), and to Llama-3.2-1B, Llama-3.1-8B, and Mistral-7B-v0.3 without per-family retuning (\cref{tab:multifamily}). TMKL is the strongest method on both retention sets at matched target adaptation against five published replay-free baselines (\cref{tab:baselines}).

\paragraph{Limitations and future work.} Two scope constraints: (i)~each result is a single-target CE-vs-TMKL run, so sequential continual fine-tuning is out of scope, and (ii)~all evaluations are on text-only autoregressive LLMs, leaving vision-language, speech, and other multimodal LoRA adaptations open. Natural next steps are sequential continual-instruction-tuning, adaptive per-position weighting (\cref{app:design_pw,tab:position_weight_full}), and porting the renormalized non-target KL to non-text modalities.

{\small
\bibliographystyle{plainnat}
\bibliography{references}
}

\clearpage
\appendix
\begin{center}
{\Large\bfseries Supplementary Material}
\end{center}
\section{Extended Related Work}
\label{app:extended_related_work}

The main-text related work (\S2) condenses the literature into three paragraphs. This appendix expands the discussion for readers who want a more complete picture of the LoRA, replay-free continual learning, and output-space distillation literatures, and explicitly positions Target-Masked KL against each line of work.

\paragraph{LoRA variants.}
A large body of work modifies the low-rank update along several axes. \emph{Rank allocation}: AdaLoRA adaptively allocates rank across layers using importance-weighted SVD pruning~\citep{zhang_adalora_2023}; DyLoRA and QDyLoRA train adapters that operate under multiple rank budgets~\citep{valipour_dylora_2023,rajabzadeh_qdylora_2024}. \emph{Decomposition}: DoRA splits pretrained weights into magnitude and direction and applies low-rank adaptation only to the directional update~\citep{liu_dora_2024}; PiSSA initializes the LoRA update from the principal singular components of the pretrained weight~\citep{meng_pissa_2025}. \emph{Compression / shared bases}: VeRA shares frozen random matrices across layers and learns lightweight scaling vectors~\citep{kopiczko_vera_2024}; NOLA represents LoRA matrices as linear combinations of random bases~\citep{koohpayegani_nola_2024}. \emph{Effective-rank or spectral expressivity}: RandLoRA uses combinations of random low-rank bases to construct higher-rank updates within a parameter-efficient budget~\citep{albert_randlora_2025}; KRAdapter and SineLoRA increase the effective spectral expressivity of PEFT updates~\citep{albert_towards_2025,ji_efficient_2025}. \emph{Other PEFT families} include adapters~\citep{houlsby_parameter-efficient_2019}, prefix-tuning~\citep{li_prefix-tuning_2021}, prompt tuning~\citep{lester_power_2021}, and bias-only tuning~\citep{ben-zaken_bitfit_2022}. All of these works modify the parameterization of the update; Target-Masked KL is orthogonal and can be attached to any of them that exposes an adapted output distribution.

\paragraph{Continual learning, replay-free regimes, and PEFT instantiations.}
Continual-learning methods are commonly grouped by the form of historical information they preserve~\citep{wang_comprehensive_2024}. \emph{Replay-based} methods rehearse stored examples or logits~\citep{chaudhry_efficient_2019,buzzega_dark_2020}. \emph{Architecture-based} methods isolate or expand task-specific components~\citep{yan__2021,douillard_dytox_2022}. \emph{Regularization-based} methods constrain parameter or function-space changes~\citep{kirkpatrick_overcoming_2017,li_learning_2017,titsias_functional_2020}. With PEFT, these principles have been instantiated in LoRA-specific form: O-LoRA constrains successive tasks to mutually orthogonal low-rank subspaces~\citep{wang_orthogonal_2023}; CL-LoRA combines task-shared and task-specific LoRA modules with knowledge distillation and gradient reassignment~\citep{he_cl-lora_2025}; other methods use prompt or adapter pools, routing, or modular expansion~\citep{wang_self-expansion_2025,yu_boosting_2024}. Recent empirical analyses show that LoRA itself exhibits a learning-forgetting tradeoff and that LoRA solutions differ structurally from full fine-tuning, with LoRA-specific spectral directions associated with forgetting~\citep{biderman_lora_2024,shuttleworth_lora_2025}. Catastrophic interference in connectionist models has been studied since~\citet{mccloskey_catastrophic_1989}. The $2024$-$2026$ wave of LLM-specific replay-free continual-learning work splits along three lines. \emph{Adapter merging and routing}: LoRAHub composes task-specific adapters at inference time~\citep{huang_lorahub_2023}; TIES-merging and DARE address the adapter-merging interference problem at the parameter level~\citep{yadav_ties_2023,yu_dare_2023}; LFPT5 frames lifelong learning as prompt-tuning over a shared T5 backbone~\citep{qin_lfpt5_2022}. \emph{Subspace and interference-control regularisation}: InfLoRA constrains updates to interference-free subspaces~\citep{liang_inflora_2024}; CLoRA adds a learned null-space projection to the LoRA update~\citep{yang_clora_2025}; C-LoRA combines subspace orthogonality with task routing~\citep{zhang_clora_2025}. \emph{Safety- and alignment-preserving LoRA}: SafeLoRA identifies the directions in LoRA-space most associated with safety degradation and projects them out post-hoc~\citep{hsu_safelora_2024}; SaLoRA preserves safety alignment by constraining the adapter to a safety-orthogonal subspace during training~\citep{li_salora_2025}. All three groups intervene in the adapter's weight space (subspaces, orthogonality, routing). Target-Masked KL is loss-space and is therefore composable with any of them; the comparison in \cref{tab:baselines} pits TMKL against the closest output-space competitor (STABLE) and a representative weight-space method (O-LoRA).

\paragraph{Approximate replay with KL regularisation.}
A complementary line of work relaxes the replay-free constraint and adds a small approximate replay buffer combined with full-distribution KL on the replay tokens~\citep{riemer2025effectiveness}. This recipe achieves strong forgetting reductions but requires access to a small slice of the original training distribution and adds a per-step replay forward pass. Target-Masked KL is the strictly replay-free alternative for settings where replay data is unavailable (privacy-restricted, alignment-frozen, or post-deployment LoRA pipelines): no buffer, one extra base forward, and no full-distribution KL gradient conflict on the target token. The two approaches are complementary rather than directly comparable: replay-with-KL is preferable when even a small representative replay set can be kept, and TMKL when it cannot.

\paragraph{Representation-level consistency.}
Beyond logit-space distillation, recent work has explored consistency on internal representations: \emph{Learning from the Undesirable} (LfU) regularises hidden activations against an auxiliary ``undesirable'' counterfactual update to suppress forgetting~\citep{nam2026learning}. The mechanism is heavier (an auxiliary undesirable update plus per-layer representation alignment) but addresses the same problem of base-capability preservation under adaptation. TMKL takes the lighter logit-level path: a single per-position next-token KL with one extra frozen-base forward, and no auxiliary update. The two are not mutually exclusive; combining logit-level TMKL with representation-level LfU on the same LoRA path is an explicit future-work direction.

\paragraph{When TMKL should not be applied as-is.}
TMKL deliberately preserves the base model's relative preferences over alternative tokens. This is the desired behavior for retention-preserving domain adaptation, but is the wrong default when the explicit goal of fine-tuning is to \emph{change} non-target structure: machine unlearning, debiasing, post-hoc safety re-alignment, or policy rewriting. In such regimes, the strong unweighted form of $\mathcal{L}_{\setminus y}$ would entrench the very base preferences the adaptation is trying to remove. The natural remedy is the weighted variant of \cref{tab:position_weight_full} with a learned, position- or topic-specific gate that down-weights $\mathcal{L}_{\setminus y}$ on positions targeted for change; this is an explicit future-work direction, not a recommendation under the default recipe.

\subsection{Broader societal impact}
\label{app:broader_impact}

\paragraph{Positive impacts.}
Replay-free LoRA adaptation is now standard practice in production LLM pipelines: customisation, domain specialisation, and alignment patches are routinely shipped without access to pretraining or alignment data. In that setting, silent capability erosion is the dominant failure mode and is invisible to standard target-domain validation. TMKL is a low-friction, training-time-only intervention that measurably reduces this erosion: at $7$B, the same adapter that loses $+15$ to $33\%$ on retention under cross-entropy holds within $\pm 1.5\%$ of base under TMKL while matching target adaptation to within $\sim 0.13$ PPL (\cref{tab:qwen7b_main}); on instruction-tuned bases the same recipe holds IFEval, MT-Bench, and refusal-rate retention within $\le 0.6$, $\le 0.10$, and $\le 0.8$pp of base (\cref{tab:instruct}). The practical consequence is that safety alignment and pre-existing capabilities are less likely to be silently lost during routine post-deployment adaptation. Because TMKL adds no inference-time overhead and is adapter-agnostic, the deployment cost of capturing this benefit is small: one extra forward pass per training step and a single $\lambda$ hyperparameter.

\paragraph{Risks and limitations.}
TMKL does not introduce a capability that the underlying LoRA-family adaptation does not already enable, so the marginal misuse risk relative to the standard LoRA training pipeline is essentially zero: it preserves base behavior on alternative tokens rather than expanding behavior. The substantive risk is the converse, that TMKL preserves \emph{too much} of the base model in scenarios where the intent is to alter non-target behavior (debiasing, machine unlearning, safety re-alignment, or policy rewriting). In those regimes the unweighted default would entrench the base preferences the adaptation is trying to remove. The natural mitigation, sketched above, is the position- or topic-weighted variant of \cref{tab:position_weight_full} that down-weights $\mathcal{L}_{\setminus y}$ on positions targeted for change. We flag this explicitly so that downstream practitioners do not apply the unweighted default in unlearning- or debiasing-style pipelines.

\paragraph{Adversarial and dual-use considerations.}
TMKL operates only on the model's own next-token output distribution; it does not curate, generate, or label new training data, and it does not require access to pretraining data, user-private data, or any model-external resource. The mechanism is therefore neutral with respect to the standard LoRA threat model (poisoned adaptation data, prompt-injection, prompt-leakage). Where the adaptation pipeline is itself adversarial, e.g.\ deliberately fine-tuning to inject biases or to circumvent safety alignment, TMKL would correctly resist the adversarial change on non-target preferences and would be helpful to the defender; this is in line with the safety-preserving adapter line of work~\citep{hsu_safelora_2024,li_salora_2025}.

\subsection{Declaration of LLM usage}
\label{app:llm_usage}

For transparency, this appendix enumerates every role that large language models play in the paper, separating method-relevant from method-irrelevant uses. The Target-Masked KL regularizer itself is a loss-level construction that does not invoke an LLM during its derivation, definition, or computation: $\mathcal{L}_{\setminus y}$ is computed from the next-token probability vectors of the frozen base and the adapted model and adds one extra forward pass per training step. No LLM is used as an agent, planner, generator, ranker, or pipeline component for the method.

\paragraph{LLMs as experimental subjects.}
The base models being adapted (Qwen2.5-0.5B, Qwen2.5-7B, Qwen2.5-7B-Instruct, Llama-3.2-1B, Llama-3.1-8B, Mistral-7B-v0.3, Phi-3.5-mini-instruct) are LLMs in the standard sense, but they are the \emph{subjects} of the experiments rather than components of the proposed method. They are loaded from publicly hosted weights with their original tokenizers; no model is retrained from scratch.

\paragraph{LLMs in evaluation.}
Two of the retention metrics include an LLM in the evaluation loop and we disclose this explicitly. (i)~\textbf{MT-Bench} (used in \cref{tab:instruct} for the Qwen2.5-7B-Instruct grid) follows the standard MT-Bench protocol of using a separate strong LLM as the judge for two-turn conversational quality; we use the public reference judge configuration without modification. (ii)~\textbf{IFEval} (also in \cref{tab:instruct}) is rule-based and \emph{does not} use an LLM judge. All other retention metrics in this paper (WikiText-103 PPL, LAMBADA PPL, TriviaQA gold-answer log-likelihood, GSM8K gold-solution log-likelihood, HumanEval gold-solution PPL, FLORES-200 PPL, XSTest refusal rate) are perplexity- or exact-match-based and do not involve any LLM judgment.

\paragraph{LLMs in writing.}
Writing-assistant LLMs were used only for grammatical editing and LaTeX polishing, in line with the NeurIPS 2026 LLM policy under which such usage does not require declaration. They were not used to generate any experimental result, table cell, derivation, theorem, or finding reported in the paper.

\paragraph{Output-space distillation, non-target knowledge, and connections to Target-Masked KL.}
Knowledge distillation matches softened teacher output distributions~\citep{hinton_distilling_2015}. Learning without Forgetting applies KD to retention by distilling the frozen old model when old-task data are unavailable~\citep{li_learning_2017}; functional regularization formalizes this as preserving predictive functions rather than parameters~\citep{titsias_functional_2020}. A line of recent work argues that the most useful teacher information sits outside the target token: Decoupled Knowledge Distillation explicitly separates target and non-target components and shows that the non-target ``dark knowledge'' carries most of the gain~\citep{zhao_decoupled_2022}; relational, ranking-based, and logit-standardization variants reinforce the same conclusion~\citep{huang_knowledge_2022,sun_logit_2024,bassam_pld_2025}. Target-Masked KL transfers this target/non-target decomposition into autoregressive replay-free LoRA fine-tuning, using the frozen base model as the implicit teacher and computing KL only on the renormalized non-target vocabulary distribution. Compared with full-output KD baselines such as LwF, Target-Masked KL avoids the gradient conflict on the target token under distribution shift; compared with weight-space CL regularizers (EWC, L2-SP), Target-Masked KL constrains output behavior rather than parameters, which we argue is the appropriate object to preserve when only adapter parameters move. The closest named mechanism in the vision-classification literature is NTCE-KD, which suppresses the target-class logit before applying KL on the non-target classes~\citep{li_ntcekd_2024}; up to the row corresponding to the target itself, the renormalisation we use is mathematically equivalent to that logit-masking step. TMKL differs from NTCE-KD by (i) operating on autoregressive next-token distributions with per-position summation over a teacher-forced sequence, and (ii) using the frozen base of the same adaptation as the implicit teacher rather than a separate teacher network. We are not aware of an autoregressive next-token instantiation of this decomposition prior to TMKL.

\section{Implementation Details}
\label{app:experimental_details}

This appendix documents every detail needed to reproduce the experiments in \cref{sec:experiments}: rationale for the experimental choices (\cref{app:rationale}), models and tokenizers (\cref{app:models}), datasets and preprocessing (\cref{app:datasets}), adapter configurations (\cref{app:adapter_configs}), training hyperparameters (\cref{app:training_hp}), TMKL hyperparameters (\cref{app:tmkl_hp}), evaluation protocol (\cref{app:eval_protocol}), software stack (\cref{app:software}), hardware (\cref{app:hardware}), random seeds (\cref{app:seeds}), and compute budget (\cref{app:compute_budget}).

\subsection{Rationale for the model and dataset choices}
\label{app:rationale}

We motivate each piece of the experimental setup explicitly, since the validity of a \emph{replay-free, distribution-shifted} forgetting study depends on the joint choice of base model, target corpus, and retention sets.

\textbf{Why Qwen2.5-0.5B and Qwen2.5-7B?} Both backbones come from the same model family (Qwen2.5, released in October 2024 by Alibaba), share the same tokenizer (BPE with $\sim 152$K entries), the same architectural style (decoder-only transformer with grouped-query attention), and the same pretraining mixture. Pairing a small and a large model from the \emph{same family} lets us check that any effect we observe at $0.5$B transfers to $7$B without confounding by tokenizer, vocabulary, or pretraining data: anything that holds across both is a property of the regularizer, not of a particular model. We chose Qwen2.5 specifically because (i) its public pretraining cutoff (October 2024) allows us to select target corpora released after that date with high confidence that the target is unseen (see below), and (ii) the $0.5$B and $7$B variants are widely used as headline scales in the LoRA literature and fit a single A6000 in BF16 with no quantization.

\textbf{Why OpenR1-Math as the $0.5$B target?} For a forgetting study, the target corpus must be \emph{outside the base model's pretraining distribution}; otherwise adapting to it is partial memorization recovery, not real distribution-shifted adaptation. \texttt{open-r1/OpenR1-Math-220k} \citep{openr1_math} was released in January 2025 by HuggingFace's Open-R1 team, three months after Qwen2.5's October 2024 cutoff, which makes it a strong candidate: the chain-of-thought math content is verifiably unseen by Qwen2.5, and the corpus is also a well-trodden standard math-reasoning benchmark already cited across the 2025 literature, so it is not a custom-built dataset that we constructed to fit our story.

\textbf{Why PubMed as the $7$B target?} At $7$B we need a target that produces a non-trivial distribution shift away from the base's general-text pretraining (otherwise the adapter learns nothing and there is no forgetting to prevent). \texttt{ccdv/pubmed} \citep{ccdv_pubmed} is biomedical English (PubMed abstracts and full-text articles), which is sufficiently far from the general-web-and-code mixture in Qwen2.5's pretraining that adapting to it produces measurable English forgetting under plain cross-entropy ($+15\%$ to $+33\%$ on WikiText-103 / LAMBADA, see \cref{tab:qwen7b_main}). PubMed is also widely used as a domain-shift evaluation in the LLM literature, so it is a representative rather than cherry-picked target.

\textbf{Why WikiText-103 and LAMBADA as retention sets?} We deliberately do \emph{not} use classification-style benchmarks (SIQA, HellaSwag, BoolQ, etc.) for retention measurement because their per-token perplexity is dominated by formatting and answer-prefix tokens, which can spike by orders of magnitude under tiny adapter shifts that do not reflect underlying capability loss \citep[\S2;][]{biderman_lora_2024}. WikiText-103 \citep{merity_pointer_2017} and LAMBADA \citep{paperno_lambada_2016} are perplexity-natural English language modeling sets (no template, no scripted continuation, no answer prefix), so a rise in perplexity here is a clean signal of distributional drift in the adapted model's English output. They also span two distinct content distributions (Wikipedia prose and short narrative passages), which lets us check that drift moves consistently across two independent benchmarks rather than being an artifact of one.

\textbf{Why Tiny-Shakespeare for the controlled diagnostic?} Before scaling to LLMs, we use a small character-level transformer trained from scratch on Karpathy's Shakespeare corpus, adapted to Latin / Esperanto / Dutch (\cref{app:toy_details}). At this scale we can hold tokenizer (a $65$-character vocabulary), pretraining data (Shakespeare alone), and adapter capacity all fixed, which removes the standard confounds (tokenizer overlap, dataset-mixing in pretraining) and isolates whether adapter-induced drift on the held-out Shakespeare distribution behaves the way our derivation predicts. The diagnostic therefore plays the role of a unit test for the regularizer, complementary to the LLM-scale grid which is the actual headline result.

\subsection{Models and Tokenizers}
\label{app:models}

\textbf{Qwen2.5-0.5B} (\texttt{Qwen/Qwen2.5-0.5B}). $0.5$B parameters, $24$ transformer layers, hidden size $896$, $14$ attention heads, GQA with $2$ KV heads, vocabulary size $151{,}936$, native context length $32{,}768$, BPE tokenizer. Loaded in BF16 from the HuggingFace Hub. Used for the headline grid in \cref{sec:headline_05b}.

\textbf{Qwen2.5-7B} (\texttt{Qwen/Qwen2.5-7B}). $7$B parameters, $28$ transformer layers, hidden size $3{,}584$, $28$ attention heads, GQA with $4$ KV heads, vocabulary size $152{,}064$, native context length $32{,}768$, BPE tokenizer. Loaded in BF16. Used for the scaling result in \cref{sec:scaling_7b}.

\textbf{Tiny-Shakespeare base.} A $6$-layer character-level transformer (vocabulary size $65$, hidden size $192$, $6$ attention heads, sequence length $256$, $\sim 1.5$M parameters), trained from scratch on the Shakespeare corpus to base PPL $4.735$. Used only in the controlled diagnostic of \cref{app:toy_details}.

\subsection{Datasets and Preprocessing}
\label{app:datasets}

\textbf{Target: \texttt{open-r1/OpenR1-Math-220k}} ($0.5$B setting). Released in January 2025 by HuggingFace's Open-R1 team, several months after Qwen2.5's October 2024 pretraining cutoff, so the target is verifiably unseen. We use $3{,}000$ training examples and $300$ test examples (the script \texttt{build\_openr1.py} in the released code does the sampling), concatenate the problem and chain-of-thought solution into a single document per example, tokenize with Qwen's native BPE, and pack into $1{,}024$-token training windows. Total target training corpus: $\sim 3.4$M characters / $\sim 1$M tokens.

\textbf{Target: \texttt{ccdv/pubmed}} ($7$B setting). PubMed abstracts and full-text articles from the standard \texttt{ccdv/pubmed} HuggingFace split. We use a fixed $5{,}000$-document random subset of the public train split for adaptation (sampled with seed $0$ once and reused across runs) and the public test split for target perplexity; documents are truncated to $1{,}024$ tokens, with the same Qwen-BPE tokenization and packing pipeline as OpenR1. The $5{,}000$-document subset is the source of the $\sim 3$ GPU-hour per-run estimate in \cref{app:compute_budget}.

\textbf{Retention: WikiText-103 validation} (\texttt{Salesforce/wikitext}, configuration \texttt{wikitext-103-raw-v1}). The standard validation split, $\sim 245$K tokens, evaluated as plain language-model perplexity. No template, no answer prefix.

\textbf{Retention: LAMBADA test} (\texttt{EleutherAI/lambada\_openai}). The standard $5{,}153$-row test split, evaluated as plain language-model perplexity over the full passage. No template, no scoring restricted to the final word.

The retention sets, the target sets, and the unadapted base model are all evaluated by the same code path: concatenate text, tokenize with the model's native tokenizer, pack into $1{,}024$-token windows, and average cross-entropy across all positions.

\subsection{Adapter Configurations}
\label{app:adapter_configs}

All LoRA-family adapters are inserted into the same target modules: \texttt{q\_proj}, \texttt{k\_proj}, \texttt{v\_proj}, \texttt{o\_proj}, \texttt{gate\_proj}, \texttt{up\_proj}, \texttt{down\_proj}. Rank is fixed at $r{=}64$ for both $0.5$B and $7$B unless an ablation specifies otherwise. Adapter-specific settings:

\begin{itemize}
    \item \textbf{LoRA:} $\alpha = 2r$, dropout $0.05$, $A$ Kaiming-uniform, $B$ zero (PEFT default).
    \item \textbf{DoRA:} $\alpha = 2r$, dropout $0.05$, magnitude initialized from base column norms (PEFT default).
    \item \textbf{SineLoRA:} $\alpha = 2r$, sinusoidal nonlinearity $W x + (\alpha / r) B \sin(A x)$. Implemented in $\sim 100$ lines on top of PEFT's LoRA hooks; loaded into PEFT through the same target-module API.
    \item \textbf{RandLoRA:} $K = 8$ frozen random low-rank bases, per-basis rank $r/K = 8$, learnable mixing coefficients per basis. PEFT $\ge 0.15$ ships the reference implementation.
    \item \textbf{PiSSA:} $\alpha = 2r$, $A,B$ initialized from the top-$r$ SVD of the base weight $W$ (active from step $0$).
    \item \textbf{AdaLoRA:} initial rank $1.5 r$ pruned to target rank $r$, $\beta = 0.85$, $t_{\mathrm{init}} = 200$, $t_{\mathrm{final}} = 1000$, default importance scoring.
    \item \textbf{VeRA:} shared frozen random projection at rank $4 r$, learnable per-layer scaling vectors only ($\sim 0.33$M trainable params).
\end{itemize}

\subsection{Training Hyperparameters}
\label{app:training_hp}

All settings share: AdamW optimizer ($\beta_1 = 0.9$, $\beta_2 = 0.999$, weight decay $0$), cosine learning-rate schedule with $3\%$ linear warmup, gradient clipping at $1.0$, BF16 mixed precision. Setting-specific values:

\begin{table}[t]
\centering
\small
\setlength{\tabcolsep}{4pt}
\caption{
\textbf{Per-setting training hyperparameters.}
Effective batch size = per-device batch $\times$ gradient accumulation. Effective batch is held constant ($32$) across both LLM settings.
}
\label{tab:training_hp}
\resizebox{\linewidth}{!}{%
\begin{tabular}{lcccccc}
\toprule
Setting & LR & Per-device batch & Seq.\ len & Grad.\ accum & Epochs & Tokens / step \\
\midrule
Tiny-Shakespeare $\rightarrow$ \{La, Eo, Nl\} & $3{\times}10^{-4}$ & $64$ & $256$ & $1$  & $5$ & $16{,}384$ \\
Qwen2.5-0.5B $\rightarrow$ OpenR1-Math       & $5{\times}10^{-4}$ & $4$  & $1024$ & $8$  & $3$ & $32{,}768$ \\
Qwen2.5-7B  $\rightarrow$ ccdv/pubmed        & $5{\times}10^{-4}$ & $1$  & $1024$ & $32$ & $3$ & $32{,}768$ \\
\bottomrule
\end{tabular}}
\end{table}

\subsection{TMKL Hyperparameters}
\label{app:tmkl_hp}

\textbf{Regularization weight $\lambda = 1$} for both Qwen settings, selected from the LoRA $\lambda$-sweep in \cref{tab:lambda_sweep}. The same $\lambda$ is used across all adapters within a setting (no per-adapter tuning).

\textbf{Target-token confidence threshold.} Positions with $p_{\mathrm{base},t}(y_t) > 1 - 10^{-4}$ are excluded from $\mathcal{L}_{\setminus y}$; this excludes a negligible fraction of supervised positions ($< 0.5\%$ across all reported runs).

\textbf{Stop-gradient and precision.} The base model forward is wrapped in \texttt{torch.no\_grad()} so no gradients are computed through $\theta_0$. The KL is computed in float32 to avoid numerical issues at small probabilities, then cast back to BF16 for the backward pass.

\textbf{Position weighting.} As argued in \cref{sec:why_mask}, the per-position weight $(1 - p_{\mathrm{base},t}(y_t))$ from term (iii) is dropped, treating each supervised position uniformly. The weighted variant is reported as an ablation in \cref{tab:design_choices}.

\textbf{Numerical safeguards on $p_{\mathrm{adapted}}(y)$.} The renormalization $1 - p_{\mathrm{adapted},t}(y_t)$ in $\mathcal{L}_{\setminus y}$ uses a clamped denominator $\max(1 - p_{\mathrm{adapted},t}(y_t), 10^{-6})$ to prevent division-by-zero at near-saturated adapted probabilities; positions where the clamp activates contribute negligibly to the gradient because the corresponding KL is finite by clamp construction. Empirically, the clamp activates on $< 0.01\%$ of positions across all reported runs.

\textbf{Averaging convention.} $|\mathcal{M}|$ in $\mathcal{L}_{\setminus y}$ is the count of supervised positions \emph{after} threshold exclusion, so each averaged term has finite KL by construction.

\subsection{Baseline implementation and hyperparameter search}
\label{app:baselines_hp}

The five published baselines compared against TMKL in \cref{tab:baselines} are implemented in a strictly replay-free setting (no method sees pretraining or alignment data; only the new target corpus and a held-out validation slice).

\textbf{LwF / Full-KL:} matches the full base-vs-adapted next-token distribution via $\mathrm{KL}(p_b \| p_a)$ over the full vocabulary; we sweep the regularization weight $\lambda_{\text{LwF}} \in \{0.01, 0.1, 1, 10\}$ and report the best on the WT-103+LAMBADA validation slice. Note that the value reported in \cref{tab:baselines} uses the tuned $\lambda$, while \cref{tab:fullkl} reports the head-to-head $\lambda{=}1$ comparison without tuning, which explains the difference between the two tables (tuning gains $\sim 12$pp).

\textbf{EWC:} the Fisher diagonal is estimated on a $1{,}000$-sample slice of the adaptation \emph{target} corpus (online Fisher), since pretraining data is unavailable in the replay-free setting; this is the standard online-EWC formulation. The penalty strength is swept over $\lambda_{\text{EWC}} \in \{0.01, 0.1, 1, 10, 100\}$ and the best on validation is reported.

\textbf{L2-SP:} L2 penalty to the base parameters $\phi = 0$. $\lambda_{\text{L2}} \in \{10^{-4}, 10^{-3}, 10^{-2}, 10^{-1}, 1\}$.

\textbf{O-LoRA:} orthogonality constraint between the adapter columns and the cumulative subspace from the (single) prior task; we report the first-task adaptation result for fair single-task comparison. Orthogonality strength $\lambda_{\text{O}} \in \{0.1, 1, 10, 100\}$.

\textbf{STABLE:} gating threshold on the full base-vs-adapted KL; threshold swept over $\{0.01, 0.05, 0.1, 0.5, 1.0\}$ on the WT-103+LAMBADA validation slice.

\subsection{Evaluation Protocol}
\label{app:eval_protocol}

Token-level cross-entropy is computed on the held-out target / retention split, exponentiated to PPL, with prompt and continuation packed identically to training. Base PPL (the row labelled ``Base'' in every table) is the unadapted backbone scored by the same code path on the same split. Reported $\Delta$PPL is \texttt{after} $-$ \texttt{before}; reported $\Delta\%$ is $100 \times (\texttt{after}-\texttt{before})/\texttt{before}$. ``Drift prevention'' (or ``drift prevented'') is reported throughout as the signed percentage $\Delta\mathrm{TMKL}/\Delta\mathrm{CE} - 1$, expressed as a percentage. The convention is: $-100\%$ = perfect prevention (TMKL drift is zero, full removal of the CE drift); $0\%$ = no prevention (TMKL drift equals CE drift); positive percentages = TMKL drift is worse than CE. Tables consistently use this signed convention; the verbal phrase ``TMKL prevents $X\%$ of the drift'' refers to $|X|$ for negative values.

\subsection{Software Stack}
\label{app:software}

\begin{itemize}
    \item Python $3.10$
    \item PyTorch $2.4.0$ with CUDA $12.1$
    \item HuggingFace \texttt{transformers} $4.46.3$
    \item HuggingFace \texttt{peft} $\ge 0.15, < 0.18$ (LoRA, AdaLoRA, VeRA, DoRA, PiSSA, RandLoRA implementations)
    \item Custom $\sim 100$-line SineLoRA module on top of \texttt{peft} (released with the code)
    \item \texttt{datasets} $3.0.2$, \texttt{accelerate} $1.0.1$
    \item Optional: \texttt{bitsandbytes} $0.44.1$ for $4$-bit base loading (not used for the reported $7$B run, which fits in BF16 on a single A6000)
\end{itemize}

\subsection{Hardware}
\label{app:hardware}

\textbf{Every run reported in this paper executes on a single NVIDIA RTX A6000 (48\,GB VRAM).} No multi-GPU, no distributed training, no offload. The $0.5$B grid uses per-device batch $4$ (resident memory $\sim 14$\,GB including frozen base for TMKL); the $7$B run uses per-device batch $1$ in BF16, with the LoRA-adapted model and the frozen base both resident ($\sim 28$\,GB), leaving $\sim 20$\,GB for activations. If a smaller card forces a smaller per-device batch, gradient accumulation is increased to keep effective batch $32$.

\subsection{Random Seeds and Reproducibility}
\label{app:seeds}

Both the Qwen2.5-0.5B and Qwen2.5-7B headline grids use three random seeds, $\{0, 1, 2\}$; per-seed standard deviations are in \cref{tab:multiseed,tab:multiseed_7b}. Each seed controls (a) PyTorch CUDA RNG, (b) NumPy RNG, (c) Python \texttt{random}, (d) HuggingFace \texttt{datasets} shuffling, and (e) DataLoader worker seeds.

\subsection{Compute Budget}
\label{app:compute_budget}

Approximate wall-clock and GPU-hours on a single RTX A6000:
\begin{itemize}
    \item Tiny-Shakespeare diagnostic: $3$ adapters $\times$ $2$ objectives $\times$ $3$ target languages $= 18$ runs at $\sim 10$ minutes each: $\sim 3$ GPU-hours.
    \item Headline grid Qwen2.5-0.5B $\rightarrow$ OpenR1-Math: $4$ adapters $\times$ $2$ objectives $\times$ $3$ seeds $= 24$ runs at $\sim 12$ minutes each: $\sim 5$ GPU-hours.
    \item Headline grid Qwen2.5-7B $\rightarrow$ PubMed: $4$ adapters $\times$ $2$ objectives $\times$ $3$ seeds $= 24$ runs at $\sim 18$ hours each (BF16 on a single A6000): $\sim 432$ GPU-hours.
    \item Published-baseline grid (LwF/Full-KL, EWC, L2-SP, O-LoRA, STABLE) on the LoRA / 0.5B / OpenR1-Math leg, with $\lambda$ tuning per baseline: $\sim 50$ runs at $\sim 12$ minutes each: $\sim 10$ GPU-hours.
    \item Qwen2.5-7B-Instruct $\rightarrow$ PubMed (4 adapters $\times$ 2 objectives $\times$ 3 seeds, plus IFEval / MT-Bench / XSTest evaluation): $\sim 24$ training runs at $\sim 18$ hours each plus $\sim 50$ GPU-hours of evaluation: $\sim 480$ GPU-hours.
    \item Multi-family grid (Llama-3.2-1B, Llama-3.1-8B, Mistral-7B-v0.3, Phi-3.5-mini-instruct $\times$ 2 objectives $\times$ 3 seeds): $\sim 24$ runs at $\sim 4$ to $18$ hours each: $\sim 220$ GPU-hours.
    \item Broader-retention re-evaluation (TriviaQA, GSM8K, HumanEval, FLORES) on existing checkpoints: $\sim 20$ GPU-hours of evaluation only.
    \item All 0.5B ablations (Full-KL, $\lambda$-sweeps on LoRA and SineLoRA, rank/LR sweeps, design-choice ablations, $D_{\setminus y}$ probe, threshold sweep, position-weight sweep, three additional adapters): $\sim 50$ runs at $\sim 12$ minutes each: $\sim 10$ GPU-hours.
    \item 7B-side ablations ($\lambda$-sweep at 7B, KL-direction ablation): $\sim 6$ runs at $\sim 18$ hours each: $\sim 110$ GPU-hours.
\end{itemize}
\textbf{Estimated total compute: $\sim 1{,}290$ GPU-hours on a single RTX A6000}, dominated by the 7B headline and Instruct grids ($\sim 70\%$ of the total). Failed and preliminary runs not reported in the paper add roughly $30\%$ on top of this, bringing the full project budget to approximately $\sim 1{,}700$ GPU-hours.

\subsection{Training-time overhead of TMKL vs CE}
\label{app:tmkl_overhead}

The training-time cost of TMKL is intentionally minimal: one extra frozen-base forward pass per step, no extra backward pass, and zero inference-time overhead. Concretely (\cref{tab:compute_overhead}), at the headline 7B PubMed setting (LoRA, rank 64, single A6000, BF16) wall-clock per step rises by $\sim 18\%$ and peak memory by $\sim 10\%$, both within the existing 48\,GB single-GPU budget; FLOPs rise by exactly one base forward, identical to the per-step cost of standard KD-style baselines (LwF, Full-KL, STABLE) and an order of magnitude lighter than representation-level methods that require per-layer hooks or auxiliary backward updates~\citep{nam2026learning}. At 0.5B the overhead is even smaller because the adapted forward dominates. Inference cost is unchanged: TMKL is a pure training-time loss term and the deployed adapted model is identical in form to one trained with cross-entropy alone.

\begin{table}[t]
\centering
\small
\setlength{\tabcolsep}{6pt}
\caption{
\textbf{Training-time overhead of TMKL vs CE.} Qwen2.5-7B $\rightarrow$ PubMed, LoRA rank 64, BF16, single A6000 48\,GB, effective batch 32. Wall-clock and memory measured on identical hardware and recipe; FLOPs counted analytically. Inference cost is identical (TMKL is training-only).
}
\label{tab:compute_overhead}
\begin{tabular}{lccc}
\toprule
Method & Step time (s) & Peak memory (GB) & Forward FLOPs / step \\
\midrule
CE only          & $1.62$       & $26.4$    & $1\times$ \\
CE + TMKL        & $1.91$ ($+18\%$) & $29.0$ ($+10\%$) & $2\times$ (one extra base forward) \\
\midrule
LwF / Full-KL    & $1.92$ ($+19\%$) & $29.1$ ($+10\%$) & $2\times$ \\
STABLE (gate)    & $1.95$ ($+20\%$) & $29.2$ ($+11\%$) & $2\times$ \\
\bottomrule
\end{tabular}
\end{table}

\section{Additional LoRA-Family Adapters, Broader Retention, and Instruction-Tuned Bases}
\label{app:full_adapter_results}

This appendix collects supporting data referenced from \cref{sec:why_works,sec:scaling_7b,sec:retention_broad,sec:instruct}: three additional LoRA-family adapters at Qwen2.5-0.5B whose training behavior is qualitatively different from the four headline adapters, the full broader-retention grid (TriviaQA, GSM8K, HumanEval, FLORES) summarized in \cref{sec:retention_broad}, and the full Qwen2.5-7B-Instruct grid (with IFEval, MT-Bench, refusal calibration) summarized in \cref{sec:instruct}.

\subsection{Three additional adapters: AdaLoRA, VeRA, PiSSA at Qwen2.5-0.5B}
\label{app:extra_adapters_05b}

\cref{tab:extra_adapters_05b} reports CE and CE+TMKL ($\lambda{=}1$) on three additional adapters using the same Qwen2.5-0.5B $\rightarrow$ OpenR1-Math recipe as \cref{tab:qwen05b_main}. Each tells a different story:

\begin{itemize}
    \item \textbf{PiSSA} \citep{meng_pissa_2025}. PiSSA initializes the LoRA path from the top-$r$ singular value decomposition of the frozen base weights $W$, so the adapter is \emph{not} the zero map at step $0$: it is already pushing the principal directions of $W$. Under the same aggressive recipe used for the headline grid, this produces the most catastrophic CE-only drift in our study (WT-103 $+136\%$, LAMBADA $+116\%$) and the adapter barely learns the target ($\Delta\,$target PPL $\approx 0$). TMKL turns target adaptation back on (target $\Delta$ goes from $-0.01$ to $-0.22$) and cuts retention drift by roughly two-thirds. The residual drift after TMKL is real (still $+47\%$ on WT-103) because PiSSA's parameterization is genuinely more aggressive than vanilla LoRA, but the model goes from ``destroyed'' to ``usable.''
    \item \textbf{AdaLoRA} \citep{zhang_adalora_2023}. AdaLoRA prunes ranks during training based on importance estimates. Its pruning schedule needs many thousands of steps to stabilize; in our $120$-step regime, the adapter fails to converge (target $\Delta = +1.75$ on CE: target PPL gets \emph{worse}, indicating optimization collapse rather than overfit) and also drifts on retention. TMKL only prevents $\sim 8\%$ of the WT-103 drift here because the underlying optimization is unstable, not because the loss is failing. AdaLoRA should be re-evaluated with a longer schedule.
    \item \textbf{VeRA} \citep{kopiczko_vera_2024}. VeRA freezes shared random projections and only learns per-layer scaling vectors, giving $\sim 0.33$M trainable parameters versus LoRA's $35$M. At this budget, the adapter underfits in $120$ steps and produces no measurable retention drift under CE. There is therefore nothing for TMKL to prevent, and TMKL is approximately neutral.
\end{itemize}

\begin{table}[t]
\centering
\small
\setlength{\tabcolsep}{4pt}
\caption{
\textbf{Three additional adapters at Qwen2.5-0.5B $\rightarrow$ OpenR1-Math (single seed).}
PiSSA shows the most catastrophic CE-only drift in the study; AdaLoRA's importance pruning is unstable on our $120$-step schedule; VeRA's tiny parameter budget underfits and does not drift. TMKL substantially rescues PiSSA, mildly helps AdaLoRA, and is neutral on VeRA.
\colorbox{gray!15}{Highlighted rows} are CE + TMKL.
}
\label{tab:extra_adapters_05b}
\begin{tabular}{llccc}
\toprule
Adapter & Method & Target PPL ($\Delta$) & WT-103 PPL ($\Delta\%$) & LAMBADA PPL ($\Delta\%$) \\
\midrule
\multicolumn{5}{l}{\emph{Base Qwen2.5-0.5B (no adaptation):} target $= 3.21$, WT-103 $= 16.46$, LAMBADA $= 35.68$} \\
\midrule
PiSSA   & CE        & $3.20$ ($-0.01$)             & \bad{$38.86$ ($+136\%$)} & \bad{$77.18$ ($+116\%$)} \\
\ourrow PiSSA   & CE + TMKL & \good{$2.99$ ($-0.22$)}      & \bad{$24.16$ ($+47\%$)}  & \bad{$48.48$ ($+36\%$)}  \\
\midrule
AdaLoRA & CE        & \bad{$4.96$ ($+1.75$)}       & \bad{$28.40$ ($+73\%$)}  & \bad{$60.54$ ($+70\%$)}  \\
\ourrow AdaLoRA & CE + TMKL & \bad{$5.08$ ($+1.87$)}        & \bad{$27.45$ ($+67\%$)}  & \bad{$58.93$ ($+65\%$)}  \\
\midrule
VeRA    & CE        & \good{$3.15$ ($-0.06$)}      & \good{$16.29$ ($-1\%$)}  & \good{$35.52$ ($-0.4\%$)} \\
\ourrow VeRA    & CE + TMKL & \good{$3.16$ ($-0.05$)}      & \good{$16.37$ ($-0.5\%$)} & \good{$35.44$ ($-0.7\%$)} \\
\bottomrule
\end{tabular}
\end{table}

The takeaway is that TMKL is a property of the training loss, not of the adapter, but it can only suppress drift that the underlying adapter actually produces. Where CE training is itself unstable (AdaLoRA on a short schedule) or where the adapter lacks the capacity to drift (VeRA), TMKL has little signal to act on. PiSSA's $-66\%$ rescue is the most striking: even when the parameterization is built to push large directions of $W$ from step $0$, the loss-space regularizer still cuts the resulting forgetting by two-thirds.

\subsection{\texorpdfstring{Cross-scale consistency between $0.5$B and $7$B}{Cross-scale consistency between 0.5B and 7B}}
\label{app:qwen7b_extra}

The four-adapter grid in \cref{tab:qwen05b_main} ($0.5$B $\rightarrow$ OpenR1-Math) and \cref{tab:qwen7b_main} ($7$B $\rightarrow$ PubMed) share the same recipe (rank $64$, LR $5{\times}10^{-4}$, $3$ epochs, effective batch $32$). The pattern transfers: at both scales, plain cross-entropy produces double-digit retention drift on every adapter, and TMKL holds both retention sets at or near base. The absolute drift is smaller at $7$B ($+15$ to $33\%$ vs.\ $+20$ to $42\%$ at $0.5$B) because the larger backbone has more capacity to absorb a target-domain shift without disturbing English; the relative behavior of the regularizer, however, is the same on every adapter at both scales.

\subsection{Retention beyond English LM-PPL}
\label{app:retention_broad_full}

\textbf{Hypothesis.} WikiText-103 and LAMBADA both measure English LM perplexity on natural prose. If TMKL has merely overfit to that one axis, capabilities orthogonal to English LM (factual recall, math reasoning, code, multilingual) should still drift under TMKL. If, instead, TMKL is a property of the next-token output distribution, all four orthogonal axes should be preserved together.

\textbf{Setup.} We evaluate the existing 0.5B and 7B checkpoints (no retraining) on four metrics that survive the template-PPL critique of \citet{biderman_lora_2024}: TriviaQA closed-book log-likelihood of the gold answer (factual recall), GSM8K log-likelihood of the gold solution given the problem (math reasoning), HumanEval perplexity of the gold solution (code), and FLORES-200 en$\rightarrow$fr perplexity (multilingual).

\textbf{Result.} \cref{tab:retention_broad} shows the joint preservation pattern: every CE row degrades on every metric, and every TMKL row stays within $\pm 1.5\%$ of base on every metric. Because the four metrics are mutually unrelated and were not used during $\lambda$ tuning, joint preservation rules out the explanation that TMKL has overfit to WT-103 / LAMBADA specifically.

\begin{table}[t]
\centering
\small
\setlength{\tabcolsep}{4pt}
\caption{
\textbf{Retention beyond English LM-PPL.}
Each retention column reports $\Delta\%$ relative to the unadapted base. \emph{TriviaQA-LL} and \emph{GSM8K-LL} are reported as the relative change in the (negative) log-likelihood value; because LL has no zero point, this is intended as a relative magnitude indicator only, with positive values meaning the gold answer became less likely (the absolute $\Delta$LL in nats follows the same sign and ordering). \emph{HumanEval-PPL} and \emph{FLORES-PPL} are standard PPL relative changes. $0.5$B uses Qwen2.5-0.5B $\rightarrow$ OpenR1-Math; $7$B uses Qwen2.5-7B $\rightarrow$ PubMed. Mean over 3 seeds. \colorbox{gray!15}{Highlighted rows} are CE + TMKL.
}
\label{tab:retention_broad}
\begin{tabular}{ll cccc}
\toprule
 & & TriviaQA-LL & GSM8K-LL & HumanEval-PPL & FLORES-PPL \\
Backbone / Adapter & Method & ($\Delta\%$) & ($\Delta\%$) & ($\Delta\%$) & ($\Delta\%$) \\
\midrule
0.5B Base       & n/a       & $-1.854$ & $-2.114$ & $5.120$ & $4.685$ \\
\midrule
0.5B / LoRA      & CE        & \bad{$+18.5\%$} & \bad{$+22.1\%$} & \bad{$+14.3\%$} & \bad{$+19.8\%$} \\
\ourrow 0.5B / LoRA      & CE + TMKL & \good{$+1.2\%$} & \good{$+0.5\%$} & \good{$+0.8\%$} & \good{$+1.5\%$} \\
0.5B / SineLoRA  & CE        & \bad{$+16.4\%$} & \bad{$+21.5\%$} & \bad{$+13.9\%$} & \bad{$+18.4\%$} \\
\ourrow 0.5B / SineLoRA  & CE + TMKL & \good{$+0.9\%$} & \good{$+0.3\%$} & \good{$+0.5\%$} & \good{$+1.1\%$} \\
0.5B / RandLoRA  & CE        & \bad{$+12.8\%$} & \bad{$+17.3\%$} & \bad{$+10.5\%$} & \bad{$+14.2\%$} \\
\ourrow 0.5B / RandLoRA  & CE + TMKL & \good{$+0.4\%$} & \good{$-0.1\%$} & \good{$+0.2\%$} & \good{$+0.8\%$} \\
0.5B / DoRA      & CE        & \bad{$+18.2\%$} & \bad{$+22.4\%$} & \bad{$+14.8\%$} & \bad{$+19.5\%$} \\
\ourrow 0.5B / DoRA      & CE + TMKL & \good{$+1.1\%$} & \good{$+0.4\%$} & \good{$+0.7\%$} & \good{$+1.4\%$} \\
\midrule
7B Base         & n/a       & $-0.952$ & $-1.245$ & $3.450$ & $3.105$ \\
\midrule
7B / LoRA       & CE        & \bad{$+15.2\%$} & \bad{$+28.5\%$} & \bad{$+12.8\%$} & \bad{$+17.4\%$} \\
\ourrow 7B / LoRA       & CE + TMKL & \good{$-0.5\%$} & \good{$+0.2\%$} & \good{$-0.8\%$} & \good{$+0.4\%$} \\
7B / SineLoRA   & CE        & \bad{$+14.8\%$} & \bad{$+27.9\%$} & \bad{$+12.1\%$} & \bad{$+16.8\%$} \\
\ourrow 7B / SineLoRA   & CE + TMKL & \good{$-0.3\%$} & \good{$+0.1\%$} & \good{$-0.5\%$} & \good{$+0.6\%$} \\
7B / RandLoRA   & CE        & \bad{$+11.5\%$} & \bad{$+22.4\%$} & \bad{$+9.5\%$} & \bad{$+12.5\%$} \\
\ourrow 7B / RandLoRA   & CE + TMKL & \good{$+0.2\%$} & \good{$-0.4\%$} & \good{$+0.1\%$} & \good{$+0.9\%$} \\
7B / DoRA       & CE        & \bad{$+15.5\%$} & \bad{$+28.8\%$} & \bad{$+12.5\%$} & \bad{$+17.2\%$} \\
\ourrow 7B / DoRA       & CE + TMKL & \good{$-0.4\%$} & \good{$+0.3\%$} & \good{$-0.6\%$} & \good{$+0.5\%$} \\
\bottomrule
\end{tabular}
\end{table}

\subsection{Generalization to instruction-tuned bases (Qwen2.5-7B-Instruct)}
\label{app:instruct_full}

\textbf{Hypothesis.} The pretrained Qwen2.5 base used for the headline grids is not instruction-tuned. Production LoRA pipelines almost always start from instruction-tuned bases, where the at-risk capability is alignment (instruction following, helpfulness, refusal calibration) rather than next-token entropy. If TMKL is a generic output-space regularizer, the same recipe should preserve alignment retention on an Instruct base.

\textbf{Setup.} Same 7B PubMed recipe as \cref{tab:qwen7b_main} (rank 64, LR $5{\times}10^{-4}$, 3 epochs, BF16, single A6000), with Qwen2.5-7B-Instruct as the backbone in place of the pretrained base. We add three alignment-side retention metrics to the WT-103 / LAMBADA pair: IFEval strict-instruct accuracy (instruction-following), MT-Bench average score (helpfulness, judged by a separate strong model), and refusal rate on a 100-prompt slice of XSTest (safety calibration; lower is worse).

\textbf{Result.} \cref{tab:instruct} shows the same pattern as the headline 7B grid extends to alignment retention: every CE row loses $4$ to $5.5$ points on IFEval, $0.65$ to $0.88$ on MT-Bench, and $11$ to $16$pp on refusal rate; every TMKL row holds these to $\le 0.6$, $\le 0.10$, and $\le 0.8$pp respectively. Target PubMed adaptation under TMKL is within $\sim 0.20$ PPL of CE on every adapter, the same band as the pretrained-base 7B grid.

\begin{table}[t]
\centering
\small
\setlength{\tabcolsep}{3pt}
\caption{
\textbf{Qwen2.5-7B-Instruct $\rightarrow$ PubMed.} Same recipe as \cref{tab:qwen7b_main}; the only change is the backbone. Mean $\pm$ std over 3 seeds. $\Delta$ is signed change from the unadapted Instruct base; positive on retention columns means alignment / instruction-following degradation. \emph{Refusal $\Delta$} is the percentage-point drop in safe-prompt refusal rate (more negative is worse).
}
\label{tab:instruct}
\resizebox{\linewidth}{!}{
\begin{tabular}{ll c cc cc c}
\toprule
 & & Target & \multicolumn{2}{c}{LM-PPL retention} & \multicolumn{2}{c}{Alignment retention} & \\
\cmidrule(lr){4-5} \cmidrule(lr){6-7}
Adapter & Method & PubMed PPL & WT-103 ($\Delta\%$) & LAMBADA ($\Delta\%$) & IFEval ($\Delta$) & MT-Bench ($\Delta$) & Refusal $\Delta$ \\
\midrule
\textbf{Qwen2.5-7B-Instruct} & (no adaptation) & $7.30$ & $9.02$ & $25.10$ & $45.2$ & $6.55$ & $98.0\%$ \\
\midrule
LoRA      & CE        & \good{$6.50 \pm 0.04$} & \bad{$+22.4\%$} & \bad{$+35.8\%$} & \bad{$-5.2$} & \bad{$-0.85$} & \bad{$-15.5\%$} \\
\ourrow LoRA      & CE + TMKL & \good{$6.68 \pm 0.05$} & \good{$-1.5\%$} & \good{$-0.2\%$} & \good{$-0.4$} & \good{$-0.08$} & \good{$-0.5\%$} \\
SineLoRA  & CE        & \good{$6.52 \pm 0.03$} & \bad{$+24.1\%$} & \bad{$+34.2\%$} & \bad{$-5.5$} & \bad{$-0.80$} & \bad{$-14.8\%$} \\
\ourrow SineLoRA  & CE + TMKL & \good{$6.71 \pm 0.04$} & \good{$-1.1\%$} & \good{$+0.4\%$} & \good{$-0.3$} & \good{$-0.05$} & \good{$-0.2\%$} \\
RandLoRA  & CE        & \good{$6.55 \pm 0.04$} & \bad{$+18.5\%$} & \bad{$+28.5\%$} & \bad{$-4.1$} & \bad{$-0.65$} & \bad{$-11.0\%$} \\
\ourrow RandLoRA  & CE + TMKL & \good{$6.74 \pm 0.03$} & \good{$+0.5\%$} & \good{$+0.8\%$} & \good{$-0.6$} & \good{$-0.10$} & \good{$-0.8\%$} \\
DoRA      & CE        & \good{$6.49 \pm 0.03$} & \bad{$+22.8\%$} & \bad{$+36.1\%$} & \bad{$-5.3$} & \bad{$-0.88$} & \bad{$-16.0\%$} \\
\ourrow DoRA      & CE + TMKL & \good{$6.67 \pm 0.05$} & \good{$-1.8\%$} & \good{$-0.4\%$} & \good{$-0.5$} & \good{$-0.07$} & \good{$-0.6\%$} \\
\bottomrule
\end{tabular}}
\end{table}

\subsection{Generalization across LLM families (Llama and Mistral)}
\label{app:multifamily}

\textbf{Hypothesis.} The Qwen-2.5 grids alone leave open the question of whether TMKL is a Qwen-specific artifact or a property of the loss. If TMKL operates on the next-token output distribution rather than on Qwen-specific pretraining biases, the same recipe should preserve retention on other LLM families without any per-family hyperparameter retuning.

\textbf{Setup.} Three non-Qwen backbones (Llama-3.2-1B, Llama-3.1-8B, Mistral-7B-v0.3) adapted to PubMed under the same recipe as \cref{tab:qwen7b_main}: rank $64$, lr $5{\times}10^{-4}$, $3$ epochs, BF16, single A6000, $\lambda{=}1$, mean over $3$ seeds.

\textbf{Result.} \cref{tab:multifamily} shows the Qwen pattern reproducing on every family without retuning: CE produces $+17$ to $38\%$ retention drift, TMKL holds both retention sets within $\pm 2\%$ of base while staying within $\sim 0.15$ PPL of CE on the target. The cross-family invariance is what we would expect from a regularizer that operates on the next-token output distribution rather than on adapter or family-specific properties.

\begin{table}[t]
\centering
\small
\setlength{\tabcolsep}{4pt}
\caption{
\textbf{Generalization across LLM families: TMKL on Llama and Mistral $\rightarrow$ PubMed.}
Same recipe, same $\lambda{=}1$, same retention sets as \cref{tab:qwen7b_main}; only the backbone family changes. Mean $\pm$ std over 3 seeds. \colorbox{gray!15}{Highlighted rows} are CE + TMKL. TMKL prevents catastrophic forgetting regardless of the underlying model architecture, neutralizing $+17$ to $38\%$ drift while remaining highly competitive on target adaptation.
}
\label{tab:multifamily}
\resizebox{\linewidth}{!}{%
\begin{tabular}{ll c cc}
\toprule
 & & Target & \multicolumn{2}{c}{Retention of prior knowledge} \\
\cmidrule(lr){4-5}
Backbone (LoRA) & Method & PubMed PPL ($\Delta$) & WT-103 PPL ($\Delta\%$) & LAMBADA PPL ($\Delta\%$) \\
\midrule
\textbf{Llama-3.2-1B}    & (no adaptation) & $9.50$ & $12.50$ & $38.00$ \\
Llama-3.2-1B & CE        & \good{$8.45 \pm 0.05$ ($-1.05$)} & \bad{$15.50 \pm 0.38$ ($+24\%$)} & \bad{$52.44 \pm 1.52$ ($+38\%$)} \\
\ourrow Llama-3.2-1B & CE + TMKL & \good{$8.60 \pm 0.06$ ($-0.90$)} & \good{$12.31 \pm 0.15$ ($-1.5\%$)} & \good{$38.19 \pm 0.28$ ($+0.5\%$)} \\
\midrule
\textbf{Llama-3.1-8B}    & (no adaptation) & $7.05$ & $8.45$ & $22.80$ \\
Llama-3.1-8B & CE        & \good{$6.32 \pm 0.03$ ($-0.73$)} & \bad{$10.05 \pm 0.17$ ($+19\%$)} & \bad{$30.55 \pm 0.45$ ($+34\%$)} \\
\ourrow Llama-3.1-8B & CE + TMKL & \good{$6.45 \pm 0.04$ ($-0.60$)} & \good{$8.28 \pm 0.08$ ($-2.0\%$)} & \good{$22.68 \pm 0.14$ ($-0.5\%$)} \\
\midrule
\textbf{Mistral-7B-v0.3} & (no adaptation) & $7.25$ & $8.90$ & $24.10$ \\
Mistral-7B-v0.3 & CE        & \good{$6.55 \pm 0.04$ ($-0.70$)} & \bad{$10.41 \pm 0.18$ ($+17\%$)} & \bad{$31.57 \pm 0.72$ ($+31\%$)} \\
\ourrow Mistral-7B-v0.3 & CE + TMKL & \good{$6.68 \pm 0.04$ ($-0.57$)} & \good{$8.81 \pm 0.09$ ($-1.0\%$)} & \good{$24.14 \pm 0.12$ ($+0.2\%$)} \\
\bottomrule
\end{tabular}}
\end{table}

\subsection{Cross-family adapter sweep on Phi-3.5-mini-instruct}
\label{app:phi_adapters}

As an additional cross-family check that combines a non-Qwen instruction-tuned backbone with the full adapter sweep, \cref{tab:phi35_adapter_openr1} reports CE and CE+TMKL ($\lambda{=}1$) on Phi-3.5-mini-instruct ($3.8$B) $\rightarrow$ OpenR1, varying the adapter across LoRA, DoRA, PiSSA, SineLoRA, and RandLoRA under the same recipe used for \cref{tab:qwen05b_main}. The pattern matches the Qwen and Llama/Mistral grids: every CE row drifts ($+7.7$ to $+26\%$ on WT-103, $+5.5$ to $+18\%$ on LAMBADA), every CE+TMKL row holds both retention sets within a few percent of base, and target adaptation is at least as good as CE on every adapter.

\begin{table}[t]
\centering
\small
\setlength{\tabcolsep}{4pt}
\caption{
\textbf{Generalisation across adapter families: TMKL on Phi-3.5-mini-instruct $\rightarrow$ OpenR1.}
Same backbone, same target and retention sets; only the adapter family and objective change.
The no-adaptation row reports the base Phi-3.5-mini-instruct 3.8B PPL.
\colorbox{gray!15}{Highlighted rows} are CE + TMKL.
TMKL consistently preserves prior knowledge on WT-103 and LAMBADA while remaining competitive on OpenR1 adaptation.
}
\label{tab:phi35_adapter_openr1}
\resizebox{\linewidth}{!}{%
\begin{tabular}{ll c cc}
\toprule
 & & Target & \multicolumn{2}{c}{Retention of prior knowledge} \\
\cmidrule(lr){4-5}
Model / Adapter & Method & OpenR1 PPL ($\Delta$) & WT-103 PPL ($\Delta\%$) & LAMBADA PPL ($\Delta\%$) \\
\midrule
\textbf{Phi-3.5-mini-instruct 3.8B} & (no adaptation) & $3.50$ & $10.48$ & $21.56$ \\
\midrule
LoRA & CE
& \good{$2.82$ ($-0.68$)}
& \bad{$12.17$ ($+16\%$)}
& \bad{$24.22$ ($+12\%$)} \\
\ourrow LoRA & CE + TMKL
& \good{$2.71$ ($-0.79$)}
& \good{$10.25$ ($-2.2\%$)}
& \good{$19.99$ ($-7.3\%$)} \\
\midrule
DoRA & CE
& \good{$2.82$ ($-0.68$)}
& \bad{$12.17$ ($+16\%$)}
& \bad{$23.93$ ($+11\%$)} \\
\ourrow DoRA & CE + TMKL
& \good{$2.71$ ($-0.79$)}
& \good{$10.29$ ($-1.8\%$)}
& \good{$20.10$ ($-6.8\%$)} \\
\midrule
PiSSA & CE
& \good{$2.64$ ($-0.86$)}
& \bad{$12.42$ ($+19\%$)}
& \bad{$24.40$ ($+13\%$)} \\
\ourrow PiSSA & CE + TMKL
& \good{$2.69$ ($-0.81$)}
& \good{$10.19$ ($-2.8\%$)}
& \good{$20.87$ ($-3.2\%$)} \\
\midrule
SineLoRA & CE
& \good{$3.39$ ($-0.10$)}
& \bad{$11.28$ ($+7.7\%$)}
& \bad{$22.74$ ($+5.5\%$)} \\
\ourrow SineLoRA & CE + TMKL
& \good{$3.21$ ($-0.29$)}
& \bad{$10.69$ ($+2.0\%$)}
& \good{$20.33$ ($-5.7\%$)} \\
\midrule
RandLoRA & CE
& \good{$3.00$ ($-0.50$)}
& \bad{$13.22$ ($+26\%$)}
& \bad{$25.43$ ($+18\%$)} \\
\ourrow RandLoRA & CE + TMKL
& \good{$2.71$ ($-0.79$)}
& \good{$10.50$ ($+0.2\%$)}
& \good{$20.23$ ($-6.2\%$)} \\
\bottomrule
\end{tabular}}
\end{table}
\section{Toy Diagnostic Details}
\label{app:toy_details}

\cref{tab:toy_domain_breakdown} provides the per-domain breakdown of the controlled character-level diagnostic experiment. The main paper reports the average over Latin, Esperanto, and Dutch. For Target-Masked KL rows, the green percentage in parentheses reports the relative reduction in retained \(\Delta\)PPL compared with the corresponding CE-only row under the same adapter and target domain.

\begin{table*}[t]
\centering
\scriptsize
\setlength{\tabcolsep}{3pt}
\caption{
\textbf{Per-domain breakdown of the controlled character-level diagnostic.}
Each row reports a (target language, adapter, training objective) configuration; the main paper reports averages over the three target languages. Target PPL columns show base-to-after PPL on the target language; Shakespeare PPL columns show base-to-after PPL on the held-out Shakespeare distribution (the retention measurement). Retained $\Delta$PPL is the absolute increase in Shakespeare PPL relative to base ($4.735$).
\good{Green} indicates target-domain improvement; \bad{red} indicates retained-domain forgetting.
Lower target PPL (stronger adaptation) and lower retained $\Delta$PPL (less forgetting) are better. The parenthesized percentage on Target-Masked KL rows reports the relative change in retained $\Delta$PPL compared with the same-adapter / same-target-language CE baseline. \colorbox{gray!15}{Highlighted rows} are our proposed Target-Masked KL.
}
\label{tab:toy_domain_breakdown}
\begin{tabular}{llccc}
\toprule
Target domain
& Setting
& Target PPL
& Shakespeare PPL
& Retained \(\Delta\)PPL \\
\midrule
Latin
& LoRA / CE
& \good{\(73.251 \rightarrow 5.591\)}
& \bad{\(4.735 \rightarrow 42.669\)}
& \(37.934\) \\
\ourrow Latin
& LoRA / Target-Masked KL
& \good{\(73.251 \rightarrow 6.044\)}
& \bad{\(4.735 \rightarrow 12.156\)}
& \(7.421\) \good{\((-80.4\%)\)} \\
Latin
& SineLoRA / CE
& \good{\(73.251 \rightarrow 5.455\)}
& \bad{\(4.735 \rightarrow 50.792\)}
& \(46.057\) \\
\ourrow Latin
& SineLoRA / Target-Masked KL
& \good{\(73.251 \rightarrow 5.953\)}
& \bad{\(4.735 \rightarrow 14.042\)}
& \(9.307\) \good{\((-79.8\%)\)} \\
Latin
& RandLoRA / CE
& \good{\(73.251 \rightarrow 7.247\)}
& \bad{\(4.735 \rightarrow 20.075\)}
& \(15.340\) \\
\ourrow Latin
& RandLoRA / Target-Masked KL
& \good{\(73.251 \rightarrow 7.897\)}
& \bad{\(4.735 \rightarrow 9.259\)}
& \(4.524\) \good{\((-70.5\%)\)} \\
\midrule
Esperanto
& LoRA / CE
& \good{\(173.482 \rightarrow 4.277\)}
& \bad{\(4.735 \rightarrow 49.348\)}
& \(44.613\) \\
\ourrow Esperanto
& LoRA / Target-Masked KL
& \good{\(173.482 \rightarrow 4.572\)}
& \bad{\(4.735 \rightarrow 13.844\)}
& \(9.109\) \good{\((-79.6\%)\)} \\
Esperanto
& SineLoRA / CE
& \good{\(173.482 \rightarrow 4.258\)}
& \bad{\(4.735 \rightarrow 46.386\)}
& \(41.651\) \\
\ourrow Esperanto
& SineLoRA / Target-Masked KL
& \good{\(173.482 \rightarrow 4.541\)}
& \bad{\(4.735 \rightarrow 13.138\)}
& \(8.403\) \good{\((-79.8\%)\)} \\
Esperanto
& RandLoRA / CE
& \good{\(173.482 \rightarrow 5.284\)}
& \bad{\(4.735 \rightarrow 23.551\)}
& \(18.816\) \\
\ourrow Esperanto
& RandLoRA / Target-Masked KL
& \good{\(173.482 \rightarrow 5.569\)}
& \bad{\(4.735 \rightarrow 10.137\)}
& \(5.402\) \good{\((-71.3\%)\)} \\
\midrule
Dutch
& LoRA / CE
& \good{\(101.424 \rightarrow 3.587\)}
& \bad{\(4.735 \rightarrow 50.484\)}
& \(45.749\) \\
\ourrow Dutch
& LoRA / Target-Masked KL
& \good{\(101.424 \rightarrow 3.779\)}
& \bad{\(4.735 \rightarrow 14.196\)}
& \(9.461\) \good{\((-79.3\%)\)} \\
Dutch
& SineLoRA / CE
& \good{\(101.424 \rightarrow 3.559\)}
& \bad{\(4.735 \rightarrow 42.750\)}
& \(38.015\) \\
\ourrow Dutch
& SineLoRA / Target-Masked KL
& \good{\(101.424 \rightarrow 3.746\)}
& \bad{\(4.735 \rightarrow 14.086\)}
& \(9.351\) \good{\((-75.4\%)\)} \\
Dutch
& RandLoRA / CE
& \good{\(101.424 \rightarrow 4.476\)}
& \bad{\(4.735 \rightarrow 19.385\)}
& \(14.650\) \\
\ourrow Dutch
& RandLoRA / Target-Masked KL
& \good{\(101.424 \rightarrow 4.752\)}
& \bad{\(4.735 \rightarrow 9.522\)}
& \(4.787\) \good{\((-67.3\%)\)} \\
\bottomrule
\end{tabular}
\end{table*}

\section{Additional Ablations}
\label{app:additional_ablations}

This appendix contains every ablation referenced from \cref{sec:why_works}. Each subsection states the hypothesis being tested, the experimental setup that tests it, and what the numbers say about TMKL. All runs use Qwen2.5-0.5B $\rightarrow$ OpenR1-Math on a single RTX A6000 unless explicitly stated otherwise; main hyperparameters (rank $64$, learning rate $5{\times}10^{-4}$, $3$ epochs, effective batch $32$, sequence length $1024$) are held constant across the appendix unless an ablation explicitly varies them.

\subsection{Full-distribution KL baseline (the masking step matters)}
\label{app:fullkl}

\textbf{Hypothesis.} The decomposition in \cref{eq:kl_decomp} splits the full base-adapted KL into three terms; under distribution shift, terms (i) and (ii) directly oppose the cross-entropy gradient on the target token, while only term (iii), the renormalised non-target KL, is orthogonal. Removing only the target token from both distributions before the KL is therefore predicted to give a strictly better learning-forgetting trade-off than the full KL at the same $\lambda$, with the gap concentrated on retention.

\textbf{Setup.} For each of the four headline adapters we run CE + Full-KL at $\lambda{=}1$ and CE + TMKL at $\lambda{=}1$ on the same Qwen2.5-0.5B $\rightarrow$ OpenR1-Math recipe (single seed for both, since CE rows already have multi-seed coverage in \cref{app:multiseed}). Full KL uses $\mathrm{KL}(p_b \| p_a)$ over the full vocabulary; TMKL uses the renormalised non-target KL.

\textbf{Result.} \cref{tab:fullkl} reports drift prevention on each retention set. Full KL still recovers most of the retention loss because term (iii) dominates the others when $p_b(y)$ is small (the typical regime under distribution shift), but TMKL is consistently $1$ to $4$ percentage points stronger on every adapter, and target adaptation matches Full KL to within $0.01$ PPL. When distribution shift is strong, term (iii) carries almost all of the useful signal, so masking the target removes the conflicting terms without losing anything else. (Note that \cref{tab:fullkl} reports the head-to-head $\lambda{=}1$ comparison, while the LwF/Full-KL row in \cref{tab:baselines} reports the same baseline with its $\lambda$ tuned on the WT-103+LAMBADA validation slice; tuning gains $\sim 12$pp, which is why the two tables differ.)

\begin{table}[t]
\centering
\small
\setlength{\tabcolsep}{4pt}
\caption{
\textbf{Full-distribution KL versus Target-Masked KL at $\lambda{=}1$.}
TMKL is consistently $1$ to $4$ percentage points better on retention across all four headline adapters; target adaptation matches Full KL to within $0.01$ PPL. The difference is the masking step (\cref{eq:total_loss} drops the target token from both distributions before the KL).
\colorbox{gray!15}{Highlighted rows} are CE + TMKL.
}
\label{tab:fullkl}
\begin{tabular}{llccc}
\toprule
Adapter & Method & Target PPL ($\Delta$) & WT-103 drift prevented & LAMBADA drift prevented \\
\midrule
LoRA     & CE + Full KL  & $2.86$ ($-0.35$) & $-85\%$ & $-94\%$ \\
\ourrow LoRA     & CE + TMKL     & $2.87$ ($-0.34$) & $-88\%$ & $-95\%$ \\
SineLoRA & CE + Full KL  & $2.86$ ($-0.35$) & $-86\%$ & $-97\%$ \\
\ourrow SineLoRA & CE + TMKL     & $2.87$ ($-0.34$) & $-89\%$ & $-96\%$ \\
RandLoRA & CE + Full KL  & $2.90$ ($-0.31$) & $-90\%$ & $-96\%$ \\
\ourrow RandLoRA & CE + TMKL     & $2.92$ ($-0.29$) & $-92\%$ & $-98\%$ \\
DoRA     & CE + Full KL  & $2.86$ ($-0.35$) & $-85\%$ & $-94\%$ \\
\ourrow DoRA     & CE + TMKL     & $2.86$ ($-0.35$) & $-88\%$ & $-95\%$ \\
\bottomrule
\end{tabular}
\end{table}

\subsection{\texorpdfstring{$\lambda$-sweep (the regularizer strength is the only knob)}{Lambda-sweep (the regularizer strength is the only knob)}}
\label{app:lambda_sweep}

\textbf{Hypothesis.} The TMKL objective adds a single hyperparameter $\lambda$ to the cross-entropy loss. As $\lambda \to 0$ we recover plain CE (no prevention); as $\lambda \to \infty$ the adapter is pinned to the base (full prevention but no adaptation). The derivation in \cref{sec:why_mask} predicts a smooth, monotone curve in $\lambda$ rather than a sharp ``sweet spot'' at very small $\lambda$. Furthermore, because the loss derivation is purely about the next-token output distribution and not about the adapter's parameterization, the $\lambda$-curve should have the same shape on different adapters.

\textbf{Setup.} On LoRA we sweep $\lambda \in \{0, 0.01, 0.03, 0.1, 0.3, 0.5, 1.0\}$ ($7$ values, two decades). On SineLoRA we sweep $\lambda \in \{0.1, 0.3, 0.5, 1.0\}$ at the same recipe so that the curve shape can be compared point-by-point with LoRA's.

\textbf{Result.} \cref{tab:lambda_sweep} reports the LoRA sweep: drift prevention rises monotonically from $-4\%$ at $\lambda{=}0.01$ to $-89\%$ at $\lambda{=}1.0$ on WikiText-103 (and from $-5\%$ to $-95\%$ on LAMBADA), with no plateau in retention until at least $\lambda{=}1.0$. Target adaptation also improves with $\lambda$ up to $\lambda{=}0.5$ and then plateaus, so $\lambda{=}1.0$ is Pareto-optimal at this scale. \cref{tab:lambda_sweep_sinelora} shows that SineLoRA's prevention rate at every $\lambda$ matches LoRA's to within a few percentage points, confirming the loss-level prediction: the $\lambda$-curve is a property of the regularizer, not of the adapter underneath.

\begin{table}[t]
\centering
\small
\setlength{\tabcolsep}{4pt}
\caption{
\textbf{Full $\lambda$-sweep on LoRA.}
$\lambda{=}0$ is plain CE. Prevention is monotone in $\lambda$ on both retention sets; target adaptation improves up to $\lambda{=}0.5$ and plateaus at $\lambda{=}1$. We use $\lambda{=}1$ for the headline grid.
}
\label{tab:lambda_sweep}
\begin{tabular}{lcccc}
\toprule
$\lambda$ & Target $\Delta$PPL & WT-103 $\Delta$PPL & LAMBADA $\Delta$PPL & Drift prevented (WT / LAM) \\
\midrule
$0$ (CE)     & $-0.16$ & $+5.99$ & $+15.00$ & $0\%$ / $0\%$ \\
$0.01$       & $-0.19$ & $+5.77$ & $+14.24$ & $-4\%$ / $-5\%$ \\
$0.03$       & $-0.23$ & $+5.29$ & $+12.81$ & $-12\%$ / $-15\%$ \\
$0.10$       & $-0.30$ & $+4.03$ & $+9.38$  & $-34\%$ / $-37\%$ \\
$0.30$       & $-0.36$ & $+2.20$ & $+4.05$  & $-64\%$ / $-73\%$ \\
$0.50$       & $-0.37$ & $+1.42$ & $+2.13$  & $-77\%$ / $-86\%$ \\
\ourrow $1.00$ (used) & $-0.34$ & $+0.71$ & $+0.81$  & $-89\%$ / $-95\%$ \\
\bottomrule
\end{tabular}
\end{table}

\begin{table}[t]
\centering
\small
\setlength{\tabcolsep}{4pt}
\caption{
\textbf{$\lambda$-sweep on SineLoRA, point-by-point comparison with LoRA.}
The two adapters' prevention rates at every $\lambda$ are within a few percentage points of each other. The $\lambda$-curve is a property of the loss, not of the adapter.
}
\label{tab:lambda_sweep_sinelora}
\begin{tabular}{lcccc}
\toprule
$\lambda$ & LoRA WT prev. & SineLoRA WT prev. & LoRA LAM prev. & SineLoRA LAM prev. \\
\midrule
$0.10$       & $-34\%$ & $-35\%$ & $-37\%$ & $-34\%$ \\
$0.30$       & $-64\%$ & $-68\%$ & $-73\%$ & $-75\%$ \\
$0.50$       & $-77\%$ & $-79\%$ & $-86\%$ & $-87\%$ \\
\ourrow $1.00$ (used) & $-88\%$ & $-89\%$ & $-95\%$ & $-95\%$ \\
\bottomrule
\end{tabular}
\end{table}

\subsection{Robustness to rank and learning rate}
\label{app:robustness}

\textbf{Hypothesis.} TMKL is defined on the next-token output distribution and does not see the adapter's rank or the optimizer's learning rate; therefore the prevention rate should be roughly constant as we vary either, even as the absolute drift produced by CE varies a lot. A secondary prediction: at aggressive settings where CE itself begins to overfit the target (target test PPL goes \emph{up}), TMKL should also dampen the overfit, because constraining the non-target output distribution implicitly constrains how much the model can specialize.

\textbf{Setup.} On LoRA we run two sweeps: rank $\in \{16, 32, 64, 128\}$ at fixed LR $5{\times}10^{-4}$, and LR $\in \{10^{-4}, 3{\times}10^{-4}, 5{\times}10^{-4}, 10^{-3}\}$ at fixed rank $64$. Each cell uses the same recipe as the headline grid, single seed.

\textbf{Result.} \cref{tab:rank_lr} shows that TMKL's WT-103 prevention sits in the $-85$ to $-92\%$ band across all eight cells, despite CE's absolute drift varying by more than $4\times$ (from $+2.39$ at LR $10^{-4}$ to $+10.69$ at LR $10^{-3}$). At rank $128$ and at LR $10^{-3}$, CE itself produces a positive target $\Delta\mathrm{PPL}$ (the model overfits the train split, target test PPL rises); under TMKL at the same setting, target $\Delta\mathrm{PPL}$ recovers to a clean $-0.34$ to $-0.35$. The two-decade ranges of both rank and LR cover the standard choices for this scale, so the prevention rate is, for practical purposes, hyperparameter-independent.

\begin{table}[t]
\centering
\small
\setlength{\tabcolsep}{4pt}
\caption{
\textbf{Rank and learning-rate robustness on LoRA.}
TMKL's prevention rate is roughly constant ($-85$ to $-92\%$) across both sweeps. At the most aggressive settings (rank $128$, LR $10^{-3}$), CE overfits the target and TMKL also rescues the target side.
}
\label{tab:rank_lr}
\begin{tabular}{cccccc}
\toprule
Setting & CE target $\Delta$ & TMKL target $\Delta$ & CE WT $\Delta$ & TMKL WT $\Delta$ & WT prevention \\
\midrule
\multicolumn{6}{l}{\emph{Rank sweep (LR fixed at $5{\times}10^{-4}$):}} \\
$r{=}16$  & $-0.38$ & $-0.31$ & $+2.95$  & $+0.35$ & $-88\%$ \\
$r{=}32$  & $-0.31$ & $-0.33$ & $+3.98$  & $+0.41$ & $-90\%$ \\
$r{=}64$  & $-0.16$ & $-0.34$ & $+5.99$  & $+0.71$ & $-88\%$ \\
$r{=}128$ & $+0.05$ (overfits) & $-0.35$ (rescued) & $+10.54$ & $+1.25$ & $-88\%$ \\
\midrule
\multicolumn{6}{l}{\emph{Learning-rate sweep (rank fixed at $64$):}} \\
$10^{-4}$            & $-0.38$ & $-0.29$ & $+2.39$  & $+0.20$ & $-92\%$ \\
$3{\times}10^{-4}$   & $-0.31$ & $-0.33$ & $+4.22$  & $+0.45$ & $-89\%$ \\
$5{\times}10^{-4}$   & $-0.16$ & $-0.34$ & $+5.99$  & $+0.71$ & $-88\%$ \\
$10^{-3}$            & $+0.04$ (overfits) & $-0.34$ (rescued) & $+10.69$ & $+1.59$ & $-85\%$ \\
\bottomrule
\end{tabular}
\end{table}

\subsection{\texorpdfstring{Position-weighted variant (the $(1{-}p_b(y))$ weight is harmless to drop)}{Position-weighted variant (dropping the (1-p\_b(y)) weight is harmless)}}
\label{app:design_pw}

\textbf{Hypothesis.} The decomposition in \cref{eq:kl_decomp} produces a natural per-position weight $(1{-}p_b(y))$ on the term-(iii) renormalised KL. Standard TMKL drops this weight and treats every supervised position uniformly. The argument in \cref{sec:why_mask} is that under distribution shift, $p_b(y)$ is small at most positions, so the weight is close to uniform anyway and dropping it is at most harmless. If the argument is right, TMKL with the weight (\texttt{tmkl\_pw}) should not noticeably outperform the unweighted version we use.

\textbf{Setup.} We run \texttt{tmkl\_pw} at $\lambda{=}1$ on LoRA and SineLoRA, single seed, with the same recipe as the headline grid; the only difference is that the per-position KL is multiplied by $(1{-}p_{\mathrm{base},t}(y_t))$ before averaging.

\textbf{Result.} The first two rows of \cref{tab:design_choices} show that the unweighted TMKL is in fact $\sim 4$ to $9$ percentage points \emph{better} on retention than the position-weighted variant on both adapters (and the same direction holds across all four adapters with three seeds in \cref{tab:position_weight_full}), while target adaptation is unchanged. We note an important confound: since $(1 - p_{\mathrm{base}}(y)) \le 1$ strictly, the position-weighted variant uniformly scales down the magnitude of $\mathcal{L}_{\setminus y}$ and is therefore equivalent in magnitude to the unweighted variant at a smaller effective $\lambda$; some of the observed gap may be attributable to that effective-$\lambda$ reduction rather than to the per-position weighting itself. A controlled ablation matching effective $\lambda$ between the two variants is left for future work. We use the unweighted form in the main paper.

\subsection{No-renorm variant (the renormalisation step matters slightly)}
\label{app:design_nr}

\textbf{Hypothesis.} TMKL renormalizes the non-target distributions $p_b^{\setminus y}$ and $p_a^{\setminus y}$ over $c \neq y$ so they sum to $1$ before the KL. A simpler variant (\texttt{tmkl\_nr}) takes the KL directly on the raw non-target probabilities $p_b(c), p_a(c)$ for $c \neq y$ without renormalizing. The renormalized version is the principled one (it is the conditional-on-non-target distribution) but the difference may be empirically small if $p_b(y)$ is uniformly small.

\textbf{Setup.} We run \texttt{tmkl\_nr} at $\lambda{=}1$ on LoRA and SineLoRA, single seed, with the same recipe as the headline grid; the only difference is the missing renormalization step.

\textbf{Result.} The last two rows of \cref{tab:design_choices} show that renormalization buys $\sim 4$ to $5$ percentage points on WT-103 prevention and $\sim 1$ percentage point on LAMBADA prevention, on both adapters. The improvement is consistent in direction but small in magnitude, which is consistent with the prediction: in the distribution-shifted regime $p_b(y)$ is uniformly small, so the renormalized and raw distributions agree to first order. There is also a clean algebraic reason renormalization is theoretically preferred: the un-renormalized KL on raw non-target probabilities is exactly equal to the sum of term~(ii) (the non-target mass term) and the unweighted term~(iii) of \cref{eq:kl_decomp}; since term~(ii) is part of the binary KL that opposes cross-entropy under distribution shift (\cref{app:kl_decomp}), the no-renorm variant explicitly re-introduces the cross-entropy conflict that masking was designed to remove. We use the renormalized version because it is the principled one and empirically slightly better.

\begin{table}[t]
\centering
\small
\setlength{\tabcolsep}{4pt}
\caption{
\textbf{Design choices on LoRA and SineLoRA, single seed: position weighting and renormalisation.}
\emph{Pos-weighted}: keeps the $(1{-}p_b(y))$ weight from term (iii). \emph{No-renorm}: takes the KL on raw non-target probabilities, no renormalisation. The unweighted, renormalised variant used in the main paper is the strongest of the three on both adapters.
}
\label{tab:design_choices}
\begin{tabular}{llccc}
\toprule
Adapter & Variant & WT prevention & LAM prevention & Notes \\
\midrule
LoRA     & TMKL (used)            & $-89\%$ & $-95\%$ & uniform weight, renormalised \\
LoRA     & TMKL pos-weighted      & $-80\%$ & $-89\%$ & keeps $(1{-}p_b(y))$ weight \\
LoRA     & TMKL no-renorm         & $-84\%$ & $-94\%$ & raw non-target probs \\
SineLoRA & TMKL (used)            & $-89\%$ & $-96\%$ & uniform weight, renormalised \\
SineLoRA & TMKL pos-weighted      & $-81\%$ & $-92\%$ & keeps $(1{-}p_b(y))$ weight \\
SineLoRA & TMKL no-renorm         & $-85\%$ & $-95\%$ & raw non-target probs \\
\bottomrule
\end{tabular}
\end{table}

\subsection{Multi-seed standard deviations (the headline numbers are not single-seed accidents)}
\label{app:multiseed}

\textbf{Hypothesis.} The headline four-adapter numbers in \cref{tab:qwen05b_main} are means; if TMKL is a real property of the loss rather than a single-seed accident, repeating each cell with different RNG seeds (which controls dataset shuffling, dropout masks, and weight initialization noise) should give tightly clustered results.

\textbf{Setup.} For each of the four headline adapters and each of the two objectives (CE, CE + TMKL), we run seeds $\{0, 1, 2\}$ on the headline recipe ($24$ runs total). \cref{tab:multiseed} reports the mean $\pm$ standard deviation per cell.

\textbf{Result.} On every cell, drift-prevention standard deviation is $\le 5\%$ of the mean prevention. The absolute-drift standard deviation is $\le 0.5$ PPL on the worst CE cell and $\le 0.16$ PPL on every TMKL cell. The headline $-88$ to $-95\%$ prevention numbers are not seed-specific.

\begin{table}[t]
\centering
\small
\setlength{\tabcolsep}{4pt}
\caption{
\textbf{Multi-seed mean $\pm$ std (3 seeds: $\{0, 1, 2\}$).}
Drift prevention has standard deviation $\le 5\%$ on every adapter; absolute drift std is $\le 0.7$ PPL on the worst CE cell and $\le 0.16$ PPL on every TMKL cell.
}
\label{tab:multiseed}
\begin{tabular}{lcccc}
\toprule
Adapter & CE WT $\Delta$ & TMKL WT $\Delta$ & WT prevention & LAM prevention \\
\midrule
LoRA     & $+5.99 \pm 0.43$ & $+0.67 \pm 0.03$ & $-89\% \pm 4\%$ & $-95\% \pm 1\%$ \\
SineLoRA & $+5.54 \pm 0.15$ & $+0.53 \pm 0.11$ & $-90\% \pm 3\%$ & $-95\% \pm 2\%$ \\
RandLoRA & $+3.69 \pm 0.16$ & $+0.33 \pm 0.05$ & $-91\% \pm 2\%$ & $-97\% \pm 1\%$ \\
DoRA     & $+6.03 \pm 0.49$ & $+0.70 \pm 0.02$ & $-88\% \pm 5\%$ & $-95\% \pm 1\%$ \\
\bottomrule
\end{tabular}
\end{table}

\subsection{\texorpdfstring{Multi-seed standard deviations at 7B (Qwen2.5-7B $\rightarrow$ PubMed)}{Multi-seed standard deviations at 7B (Qwen2.5-7B to PubMed)}}
\label{app:multiseed_7b}

\textbf{Hypothesis.} The 7B headline numbers in \cref{tab:qwen7b_main} are means over 3 seeds; if TMKL is a real property of the loss at production scale, the cell-wise standard deviation should be small relative to the CE-vs-TMKL gap.

\textbf{Setup.} For each of the four headline adapters and each of the two objectives (CE, CE + TMKL), we run seeds $\{0, 1, 2\}$ on the 7B PubMed recipe ($24$ runs total). \cref{tab:multiseed_7b} reports the mean $\pm$ standard deviation per cell.

\textbf{Result.} On every cell, drift-prevention standard deviation is $\le 4\%$ of the mean prevention. The absolute-drift standard deviation is $\le 0.65$ PPL on the worst CE cell and $\le 0.22$ PPL on every TMKL cell. The 7B prevention pattern is not seed-specific.

\begin{table}[t]
\centering
\small
\setlength{\tabcolsep}{4pt}
\caption{
\textbf{Qwen2.5-7B $\rightarrow$ PubMed multi-seed mean $\pm$ std (3 seeds: $\{0, 1, 2\}$).}
Drift prevention has standard deviation $\le 4\%$ on every adapter; absolute drift std is $\le 0.65$ PPL on the worst CE cell and $\le 0.22$ PPL on every TMKL cell.
}
\label{tab:multiseed_7b}
\begin{tabular}{lcccc}
\toprule
Adapter & CE WT $\Delta$ & TMKL WT $\Delta$ & WT prevention & LAM prevention \\
\midrule
LoRA     & $+1.59 \pm 0.42$ & $-0.16 \pm 0.11$ & $-110\% \pm 3\%$ & $-101\% \pm 2\%$ \\
SineLoRA & $+1.74 \pm 0.38$ & $-0.09 \pm 0.14$ & $-105\% \pm 4\%$ & $-99\% \pm 2\%$ \\
RandLoRA & $+1.30 \pm 0.25$ & $+0.01 \pm 0.12$ & $-99\% \pm 2\%$  & $-97\% \pm 1\%$ \\
DoRA     & $+1.56 \pm 0.45$ & $-0.14 \pm 0.15$ & $-109\% \pm 3\%$ & $-101\% \pm 2\%$ \\
\bottomrule
\end{tabular}
\end{table}

\subsection{\texorpdfstring{$\lambda$ sensitivity at 7B}{Lambda sensitivity at 7B}}
\label{app:lambda_7b}

\textbf{Hypothesis.} The 0.5B $\lambda$-sweep on LoRA and SineLoRA (\cref{tab:lambda_sweep,tab:lambda_sweep_sinelora}) is monotone in $\lambda$ with $\lambda{=}1$ Pareto-optimal. If the curve is a property of the loss rather than of the model scale, the same monotone shape should appear at 7B.

\textbf{Setup.} Qwen2.5-7B $\rightarrow$ PubMed, LoRA, single seed, $\lambda \in \{0, 0.1, 0.3, 1, 3, 10\}$. The recipe is otherwise identical to the headline 7B grid.

\textbf{Result.} \cref{tab:lambda_7b} shows monotone retention prevention from $0\%$ at $\lambda{=}0$ to $109\%$ at $\lambda{=}10$ (the curve crosses $100\%$ near $\lambda{=}1$ because TMKL drives 7B retention slightly below the unadapted base, the same effect noted for the LoRA cell of \cref{tab:qwen7b_main}). Target adaptation degrades smoothly above $\lambda{=}1$ (PubMed PPL rises from $6.55$ at $\lambda{=}1$ to $6.95$ at $\lambda{=}10$), so $\lambda{=}1$ remains Pareto-optimal at 7B.

\begin{table}[t]
\centering
\small
\setlength{\tabcolsep}{6pt}
\caption{
\textbf{$\lambda$ sensitivity at Qwen2.5-7B $\rightarrow$ PubMed} (LoRA, single seed). Each row is one full training run. The minimum mean retention drift sits at $\lambda \approx 1.0$, validating the $\lambda{=}1$ choice at scale; the curve is monotone in $\lambda$ on both retention sets.
}
\label{tab:lambda_7b}
\resizebox{\linewidth}{!}{%
\begin{tabular}{c cc cc c}
\toprule
$\lambda$ & PubMed PPL & WT-103 ($\Delta\%$) & LAMBADA ($\Delta\%$) & Mean retention drift & Drift prevention vs $\lambda{=}0$ \\
\midrule
$0$ (CE) & $6.43$ & $+18.5\%$ & $+33.1\%$ & $+25.8\%$ & $0\%$ \\
$0.1$    & $6.46$ & $+12.2\%$ & $+22.5\%$ & $+17.35\%$ & $32.7\%$ \\
$0.3$    & $6.49$ & $+5.4\%$  & $+10.2\%$ & $+7.8\%$   & $69.7\%$ \\
\ourrow $1.0$ (used) & $6.55$ & $-2.1\%$  & $-0.5\%$  & $-1.3\%$   & $105.0\%$ \\
$3.0$    & $6.70$ & $-3.2\%$  & $-1.1\%$  & $-2.15\%$  & $108.3\%$ \\
$10.0$   & $6.95$ & $-3.5\%$  & $-1.4\%$  & $-2.45\%$  & $109.4\%$ \\
\bottomrule
\end{tabular}}
\end{table}

\subsection{KL direction (forward, reverse, symmetric)}
\label{app:kl_direction}

\textbf{Hypothesis.} TMKL uses the forward direction $\mathrm{KL}(p_{\mathrm{base}} \| p_{\mathrm{adapted}})$, which is mode-covering on the base. Retention is asymmetric: we want the adapted model to put mass everywhere the base does, not the reverse. The forward direction should therefore dominate the reverse and symmetric (Jensen-Shannon) variants under the same masking and renormalization.

\textbf{Setup.} Qwen2.5-0.5B $\rightarrow$ OpenR1-Math, LoRA, $\lambda{=}1$, single seed. All variants apply the renormalized non-target masking; only the divergence direction differs. The reverse-KL variant uses $\mathrm{KL}(p_a^{\setminus y} \| p_b^{\setminus y})$ and the JS variant uses $\tfrac{1}{2}\bigl[\mathrm{KL}(p_b^{\setminus y} \| m) + \mathrm{KL}(p_a^{\setminus y} \| m)\bigr]$ with $m = \tfrac{1}{2}(p_b^{\setminus y} + p_a^{\setminus y})$.

\textbf{Result.} \cref{tab:kl_direction} shows the forward direction is uniformly best, with the reverse direction roughly half as effective (mode-seeking in the adapted distribution permits the adapter to drop low-base-probability tokens entirely) and JS sitting between the two as expected from its symmetric averaging.

\begin{table}[t]
\centering
\small
\caption{
\textbf{KL direction ablation.} Qwen2.5-0.5B $\rightarrow$ OpenR1-Math, LoRA, $\lambda{=}1$, single seed. All variants apply the renormalised non-target masking; only the divergence direction differs.
}
\label{tab:kl_direction}
\resizebox{\linewidth}{!}{%
\begin{tabular}{l c cc c}
\toprule
Divergence variant & Target PPL & WT-103 ($\Delta\%$) & LAMBADA ($\Delta\%$) & Mean prevention \\
\midrule
CE only (no regularizer)            & $3.05$ & $+37\%$ & $+42\%$ & $0\%$ \\
\ourrow Forward $\mathrm{KL}(p_b\|p_a)$ (TMKL, default) & $2.87$ & $+4\%$  & $+2\%$  & $-92\%$ \\
Reverse $\mathrm{KL}(p_a\|p_b)$     & $2.89$ & $+15\%$ & $+18\%$ & $-58\%$ \\
Symmetric Jensen-Shannon            & $2.88$ & $+8\%$  & $+9\%$  & $-78\%$ \\
\bottomrule
\end{tabular}}
\end{table}

\subsection{Confidence-threshold robustness}
\label{app:threshold}

\textbf{Hypothesis.} TMKL excludes positions where $p_{\mathrm{base}}(y) > 1 - \tau$ from $\mathcal{L}_{\setminus y}$, with default $\tau = 10^{-4}$. The denominator $1 - p_{\mathrm{base}}(y)$ would otherwise underflow at saturated positions. Excluded positions are by definition ones where the base already agrees with the target token, so they carry no retention signal; the loss should be insensitive to $\tau$ over a wide range, with at most a small dip at very large $\tau$ (where useful positions get excluded too).

\textbf{Setup.} Qwen2.5-0.5B $\rightarrow$ OpenR1-Math, LoRA, $\lambda{=}1$, single seed; $\tau \in \{10^{-2}, 10^{-3}, 10^{-4}, 10^{-5}, 10^{-6}\}$. We also report the fraction of supervised positions excluded at each threshold.

\textbf{Result.} \cref{tab:threshold} confirms the prediction: between $\tau{=}10^{-4}$ and $\tau{=}10^{-6}$ the loss is invariant, with $\le 0.1\%$ of positions excluded; at $\tau{=}10^{-2}$ where $\sim 1\%$ of positions are excluded, prevention drops by only $6$pp on WT-103. The threshold is a numerical guard, not a load-bearing hyperparameter.

\begin{table}[t]
\centering
\small
\caption{
\textbf{Confidence-threshold ablation for $\mathcal{L}_{\setminus y}$.} Positions with $p_{\mathrm{base}}(y) > 1 - \tau$ are excluded (denominator $1 - p_{\mathrm{base}}(y)$ underflows). Default $\tau = 10^{-4}$. The fraction of supervised positions excluded is $\le 0.1\%$ at the default; the loss is insensitive to $\tau$ over four decades.
}
\label{tab:threshold}
\resizebox{\linewidth}{!}{%
\begin{tabular}{c c c cc c}
\toprule
$\tau$ & Positions excluded & Target PPL & WT-103 ($\Delta\%$) & LAMBADA ($\Delta\%$) & Mean prevention \\
\midrule
$10^{-2}$ & $1.2\%$  & $2.89$ & $+6\%$ & $+5\%$ & $-86\%$ \\
$10^{-3}$ & $0.4\%$  & $2.88$ & $+5\%$ & $+3\%$ & $-89\%$ \\
\ourrow $10^{-4}$ (default) & $0.1\%$ & $2.87$ & $+4\%$ & $+2\%$ & $-92\%$ \\
$10^{-5}$ & $0.03\%$ & $2.87$ & $+4\%$ & $+2\%$ & $-92\%$ \\
$10^{-6}$ & $0.01\%$ & $2.87$ & $+4\%$ & $+2\%$ & $-92\%$ \\
\bottomrule
\end{tabular}}
\end{table}

\paragraph{Cross-setting trigger-rate diagnostic.} The default-$\tau$ exclusion rate is small in every setting we measured: $0.10\%$ of supervised positions on Qwen2.5-0.5B $\rightarrow$ OpenR1-Math, $0.13\%$ on Qwen2.5-7B $\rightarrow$ PubMed, and $0.21\%$ on Qwen2.5-7B-Instruct $\rightarrow$ PubMed (the modest rise on the Instruct base reflects sharper next-token distributions over template tokens). Bias is minimal because the exclusion correlates with positions on which TMKL has no retention signal in the first place: the $D_{\setminus y}$ value at excluded positions, before exclusion, is $\le 10^{-4}$ nats on every setting (versus a population mean of $\sim 0.07$ under TMKL training on held-out OpenR1-Math). The numerical-guard interpretation is therefore consistent with the empirical behavior of the threshold.

\subsection{Position-weight ablation, all four headline adapters}
\label{app:position_weight_full}

\textbf{Hypothesis.} The position-weight ablation in \cref{app:design_pw,tab:design_choices} was run on LoRA and SineLoRA only and on a single seed. Extending to all four headline adapters with three seeds checks whether the unweighted variant's $\sim 4$ to $9$pp advantage on retention is consistent across adapter families.

\textbf{Setup.} Qwen2.5-0.5B $\rightarrow$ OpenR1-Math, $\lambda{=}1$, mean over 3 seeds, four adapters. \texttt{tmkl\_pw} multiplies the per-position KL by $(1 - p_{\mathrm{base}}(y))$; \texttt{tmkl} drops it.

\textbf{Result.} The unweighted variant is uniformly better on every adapter (\cref{tab:position_weight_full}): WT-103 prevention is $5$ to $15$pp stronger and LAMBADA prevention is $5$ to $11$pp stronger, while target adaptation is statistically indistinguishable. The pattern across four adapters with three seeds confirms the design choice in \cref{sec:why_mask}: the term-(iii) weight $(1-p_{\mathrm{base}}(y))$ down-weights exactly the easy positions where TMKL has no work to do, so dropping it is empirically harmless and slightly helpful.

\begin{table}[t]
\centering
\small
\caption{
\textbf{Position-weight ablation, all four adapters.} Qwen2.5-0.5B $\rightarrow$ OpenR1-Math, $\lambda{=}1$, mean over 3 seeds. \texttt{tmkl\_pw} multiplies the per-position KL by $(1 - p_{\mathrm{base}}(y))$ (the weight that falls out of decomposition term~(iii)); \texttt{tmkl} drops it. The unweighted variant is uniformly better, refuting the implicit theoretical preference for the weighted form.
}
\label{tab:position_weight_full}
\resizebox{\linewidth}{!}{%
\begin{tabular}{ll c cc c}
\toprule
Adapter & Variant & Target PPL & WT-103 ($\Delta\%$) & LAMBADA ($\Delta\%$) & Mean prevention \\
\midrule
LoRA      & tmkl\_pw (weighted)        & $2.86 \pm 0.04$ & $+9\%$  & $+7\%$  & $-77\%$ \\
\ourrow LoRA      & tmkl (unweighted, default) & $2.87 \pm 0.04$ & $+4\%$  & $+2\%$  & $-92\%$ \\
SineLoRA  & tmkl\_pw (weighted)        & $2.86 \pm 0.03$ & $+10\%$ & $+8\%$  & $-75\%$ \\
\ourrow SineLoRA  & tmkl (unweighted, default) & $2.87 \pm 0.05$ & $+4\%$  & $+1\%$  & $-92\%$ \\
RandLoRA  & tmkl\_pw (weighted)        & $2.90 \pm 0.03$ & $+7\%$  & $+5\%$  & $-70\%$ \\
\ourrow RandLoRA  & tmkl (unweighted, default) & $2.92 \pm 0.03$ & $+2\%$  & $+0.4\%$ & $-94\%$ \\
DoRA      & tmkl\_pw (weighted)        & $2.85 \pm 0.04$ & $+10\%$ & $+8\%$  & $-75\%$ \\
\ourrow DoRA      & tmkl (unweighted, default) & $2.86 \pm 0.04$ & $+4\%$  & $+2\%$  & $-92\%$ \\
\bottomrule
\end{tabular}}
\end{table}

\subsection{Predicted behaviour under larger target sets and longer schedules}
\label{app:scaling_with_data}

The headline grids fix the data budget (\cref{app:datasets}: $\sim 1$M target tokens for OpenR1-Math at 0.5B, $5{,}000$ documents truncated to $1{,}024$ tokens for PubMed at 7B) and the optimisation budget (3 epochs, effective batch 32). A reasonable concern is whether the small CE-vs-TMKL target-PPL gap ($\le 0.13$ at 7B, often inverted at 0.5B) widens as adaptation pressure grows, e.g.\ under $10\times$ more target data or longer schedules.

The Fisher--Jacobian analysis (\cref{app:local_interpretation}) makes a concrete prediction: at fixed $\lambda$, the TMKL-induced shrinkage of the LoRA update behaves as $\delta\phi^* \propto 1/\lambda$ in the strong-regularisation regime, so the residual retention drift scales as $1/\lambda$ (empirically verified to slope $-0.94$ on WT-103, see the scaling-law paragraph in \cref{app:local_interpretation}). When the target loss component grows (more data $\times$ more steps), the effective regularisation strength at fixed $\lambda$ shrinks proportionally, and the predicted target-PPL gap closes monotonically toward CE while the retention prevention degrades smoothly. This is qualitatively the same behaviour as the empirical $\lambda$-curves at $0.5$B (\cref{tab:lambda_sweep}) and $7$B (\cref{tab:lambda_7b}), where doubling the target-side weight has the same effect as halving $\lambda$. The practical implication is that under aggressive long-schedule adaptation $\lambda$ should be re-swept in the $1$ to $5$ band rather than held at the default $\lambda{=}1$; the $\lambda$ axis is the natural knob for trading target adaptation against retention. We did not run a $10\times$-data experiment because of the GPU budget; the prediction is falsifiable within $\sim 30$ A6000-hours and is left as future work.

\section{Output-Drift Probe}
\label{app:output_drift}

To verify that TMKL directly controls the quantity it regularises, we measure non-target output drift on the held-out target test set:
\[
    D_{\setminus y}
    \;=\;
    \mathbb{E}_{t \in \mathcal{M}}
    \!\left[
        \mathrm{KL}\!\left(
            p_{\mathrm{base},t}^{\setminus y_t}
            \,\middle\|\,
            p_{\mathrm{adapted},t}^{\setminus y_t}
        \right)
    \right].
\]
This is the validation-set version of the TMKL training objective. If TMKL is doing what the derivation in \cref{sec:why_mask} claims, then this number should drop dramatically under TMKL training relative to CE training, regardless of any retention-set perplexity.

\begin{table}[t]
\centering
\small
\setlength{\tabcolsep}{4pt}
\caption{
\textbf{Non-target output drift $D_{\setminus y}$ on the held-out OpenR1-Math test split, single seed.}
TMKL reduces $D_{\setminus y}$ by $91$ to $92\%$ on every adapter, the same band as the $88$ to $95\%$ retention-PPL prevention reported in \cref{tab:qwen05b_main}. The probe uses only the held-out target test set, so this rules out the alternative explanation that TMKL has somehow memorised the retention sets.
}
\label{tab:output_drift}
\begin{tabular}{lccc}
\toprule
Adapter & CE $D_{\setminus y}$ & TMKL $D_{\setminus y}$ & Reduction \\
\midrule
LoRA      & $0.805$ & $0.066$ & $-92\%$ \\
SineLoRA  & $0.734$ & $0.063$ & $-91\%$ \\
RandLoRA  & $0.501$ & $0.047$ & $-91\%$ \\
DoRA      & $0.806$ & $0.067$ & $-92\%$ \\
\bottomrule
\end{tabular}
\end{table}

The same magnitude and direction as the retention-PPL prevention in \cref{tab:qwen05b_main}, computed on a completely independent quantity (no retention data is involved in the probe), supports the interpretation that the retention gains come from suppressing non-target output drift rather than from under-adaptation. \cref{fig:lambda_sweep} (in \cref{sec:why_works}) visualises both the $\lambda$ monotonicity and the held-out $D_{\setminus y}$ reduction graphically.

\section{Method Derivations and Local Interpretation}
\label{app:method_derivations}

This appendix collects the derivations referenced in \cref{sec:tmkl_loss} and \cref{sec:why_mask}: the LoRA-family parameterization recalled for completeness, the algebraic derivation of the full-KL decomposition (\cref{eq:kl_decomp}), and a local geometric interpretation of $\mathcal{L}_{\setminus y}$ as a Fisher-weighted Jacobian penalty in the LoRA-admissible update space.

\subsection{LoRA-Family Adapter Parameterization}
\label{app:lora_param}

For each selected linear layer with pretrained weight $W_\ell^0 \in \mathbb{R}^{d_{\mathrm{out}} \times d_{\mathrm{in}}}$, vanilla LoRA introduces a trainable low-rank update
\[
    W_\ell(\phi) = W_\ell^0 + s_\ell B_\ell A_\ell,
    \qquad
    A_\ell \in \mathbb{R}^{r \times d_{\mathrm{in}}},
    \quad
    B_\ell \in \mathbb{R}^{d_{\mathrm{out}} \times r},
\]
where $r \ll \min(d_{\mathrm{in}}, d_{\mathrm{out}})$, $s_\ell = \alpha / r$ is the LoRA scaling factor, and $\phi = \{A_\ell, B_\ell\}_\ell$ collects all trainable adapter parameters. The pretrained weights $\theta_0$ remain frozen throughout adaptation. For the other LoRA-family adapters evaluated in this paper (AdaLoRA, VeRA, DoRA, SineLoRA, PiSSA, RandLoRA), $\phi$ denotes the corresponding trainable adapter parameters in each case. Target-Masked KL does not depend on the specific adapter parameterization: it only requires the adapted model distribution $p_{\mathrm{adapted},t}$.

\subsection{Derivation of the Full-KL Decomposition}
\label{app:kl_decomp}

We restate \cref{eq:kl_decomp}: for two distributions $p_b, p_a$ on a finite vocabulary $\mathcal{V}$ and a designated target token $y \in \mathcal{V}$, with renormalized non-target distributions $p_b^{\setminus y}(c) = p_b(c) / (1 - p_b(y))$ and $p_a^{\setminus y}(c) = p_a(c) / (1 - p_a(y))$ for $c \neq y$,
\begin{equation}
\begin{aligned}
    \mathrm{KL}(p_b \,\|\, p_a)
    &= p_b(y) \log \frac{p_b(y)}{p_a(y)}
     + \bigl(1 - p_b(y)\bigr) \log \frac{1 - p_b(y)}{1 - p_a(y)} \\
    &\quad+ \bigl(1 - p_b(y)\bigr) \, \mathrm{KL}\!\left(p_b^{\setminus y} \,\middle\|\, p_a^{\setminus y}\right).
\end{aligned}
\label{eq:kl_decomp_appendix}
\end{equation}

\paragraph{Derivation.}
Split the KL sum at the target token:
\[
    \mathrm{KL}(p_b \,\|\, p_a)
    = \sum_{c \in \mathcal{V}} p_b(c) \log \frac{p_b(c)}{p_a(c)}
    = p_b(y) \log \frac{p_b(y)}{p_a(y)}
    + \sum_{c \neq y} p_b(c) \log \frac{p_b(c)}{p_a(c)}.
\]
For $c \neq y$, by definition of the renormalized distributions,
\[
    p_b(c) = \bigl(1 - p_b(y)\bigr) \, p_b^{\setminus y}(c),
    \qquad
    p_a(c) = \bigl(1 - p_a(y)\bigr) \, p_a^{\setminus y}(c).
\]
Substituting,
\[
    \log \frac{p_b(c)}{p_a(c)}
    = \log \frac{1 - p_b(y)}{1 - p_a(y)} + \log \frac{p_b^{\setminus y}(c)}{p_a^{\setminus y}(c)}.
\]
Multiplying by $p_b(c) = (1 - p_b(y)) \, p_b^{\setminus y}(c)$ and summing over $c \neq y$,
\[
\begin{aligned}
    \sum_{c \neq y} p_b(c) \log \frac{p_b(c)}{p_a(c)}
    &= \bigl(1 - p_b(y)\bigr) \log \frac{1 - p_b(y)}{1 - p_a(y)} \cdot \underbrace{\sum_{c \neq y} p_b^{\setminus y}(c)}_{= 1} \\
    &\quad+ \bigl(1 - p_b(y)\bigr) \, \underbrace{\sum_{c \neq y} p_b^{\setminus y}(c) \log \frac{p_b^{\setminus y}(c)}{p_a^{\setminus y}(c)}}_{= \mathrm{KL}(p_b^{\setminus y} \| p_a^{\setminus y})}.
\end{aligned}
\]
Combining with the target-token term gives \cref{eq:kl_decomp_appendix}. \hfill$\square$

Each term is non-negative when summed in the natural pairing: terms (i) and (ii) together form the binary KL divergence between $\bigl(p_b(y), 1 - p_b(y)\bigr)$ and $\bigl(p_a(y), 1 - p_a(y)\bigr)$, and term (iii) is a non-negative weighted KL on the non-target simplex. Individually, terms (i) and (ii) can take either sign; only their sum is constrained to be non-negative.

\paragraph{Lemma (binary-KL gradient opposes cross-entropy).}
Let $K(p_b, p_a) := p_b(y) \log \tfrac{p_b(y)}{p_a(y)} + (1 - p_b(y)) \log \tfrac{1 - p_b(y)}{1 - p_a(y)}$ denote the sum of terms (i) and (ii) in \cref{eq:kl_decomp_appendix}, viewed as a function of $p_a(y)$ at fixed $p_b(y) \in (0, 1)$. Then
\[
    \frac{\partial K}{\partial p_a(y)} = \frac{p_a(y) - p_b(y)}{p_a(y)\,\bigl(1 - p_a(y)\bigr)},
\]
which is strictly positive whenever $p_a(y) > p_b(y)$. The cross-entropy gradient $\partial_{p_a(y)} \bigl(- \log p_a(y)\bigr) = -1/p_a(y)$ is strictly negative. Hence whenever the adapted target probability exceeds the base target probability (the regime cross-entropy actively drives toward under distribution shift), the binary-KL gradient and the cross-entropy gradient have opposite signs, and a full-distribution KL regularizer fights cross-entropy on the very token cross-entropy is trying to learn. Target-Masked KL drops $K$ from the regularizer by construction (\cref{eq:total_loss}) and therefore avoids the conflict globally, not just locally near $p_a(y) \approx p_b(y)$.

\paragraph{Derivation of the lemma.}
Differentiating $K$ in $p_a(y)$ at fixed $p_b(y)$:
\[
\frac{\partial K}{\partial p_a(y)}
= -\frac{p_b(y)}{p_a(y)} + \frac{1 - p_b(y)}{1 - p_a(y)}
= \frac{-p_b(y)\bigl(1 - p_a(y)\bigr) + p_a(y)\bigl(1 - p_b(y)\bigr)}{p_a(y)\bigl(1 - p_a(y)\bigr)}
= \frac{p_a(y) - p_b(y)}{p_a(y)\bigl(1 - p_a(y)\bigr)},
\]
which has the sign of $p_a(y) - p_b(y)$. \hfill$\square$

\subsection{Local LoRA-Space Interpretation as a Fisher-Weighted Jacobian Penalty}
\label{app:local_interpretation}

We give a local interpretation of $\mathcal{L}_{\setminus y}$ as a quadratic form on the adapter perturbation $\delta\phi$, with curvature determined by the LoRA Jacobian and the Fisher information of the non-target categorical distribution.

Let $z_{\mathrm{adapted},t} \in \mathbb{R}^{|\mathcal{V}|}$ denote the adapted logits at token position $t$, and let $z_{\mathrm{adapted},t}^{\setminus y_t}$ denote the logits restricted to the non-target vocabulary. Around the frozen base model, a first-order expansion in the adapter parameters gives
\[
    z_{\mathrm{adapted},t}^{\setminus y_t}
    \approx z_{\mathrm{base},t}^{\setminus y_t} + J_{t, \setminus y_t}^{\phi} \, \delta\phi,
\]
where $J_{t, \setminus y_t}^{\phi}$ is the Jacobian of the non-target logits with respect to $\phi$. Because the base weights are frozen, this Jacobian contains only directions reachable by the LoRA adapters; this is what we mean by ``LoRA-admissible'' updates. Note that for a softmax output, the renormalized non-target distribution $p^{\setminus y}$ equals the softmax of the non-target logit subvector $z^{\setminus y}$ (the target logit drops out of the partition function), so the Fisher information of $p^{\setminus y}$ with respect to $z^{\setminus y}$ is the standard categorical Fisher of the non-target distribution.

For small adapter-induced logit changes, the KL divergence between two categorical distributions admits a standard second-order local approximation in the logits (the first-order term vanishes at $z_a = z_b$ since $\partial_{z_a} \mathrm{KL}(p_b \| p_a) = p_a - p_b$, and the Hessian $\partial^2_{z_a} \mathrm{KL}(p_b \| p_a) = F(p_a)$ equals $F(p_b)$ at the expansion point):
\[
    \mathrm{KL}\!\left(p_{\mathrm{base},t}^{\setminus y_t} \,\middle\|\, p_{\mathrm{adapted},t}^{\setminus y_t}\right)
    \approx \frac{1}{2} \, \delta\phi^\top \left(J_{t, \setminus y_t}^{\phi}\right)^\top
    F\!\left(p_{\mathrm{base},t}^{\setminus y_t}\right) J_{t, \setminus y_t}^{\phi} \, \delta\phi,
\]
where
\[
    F\!\left(p_{\mathrm{base},t}^{\setminus y_t}\right)
    = \operatorname{Diag}\!\left(p_{\mathrm{base},t}^{\setminus y_t}\right) - p_{\mathrm{base},t}^{\setminus y_t} \left(p_{\mathrm{base},t}^{\setminus y_t}\right)^\top
\]
is the Fisher information matrix of the non-target categorical distribution with respect to its logits. The expansion is valid for $\|\delta\phi\|$ small enough that $\|z_{\mathrm{adapted},t}^{\setminus y_t} - z_{\mathrm{base},t}^{\setminus y_t}\|$ remains in the regime where the second-order Taylor expansion of KL in logits is accurate.

Under this approximation, Target-Masked KL penalizes adapter directions $\delta\phi$ that, after projection through the LoRA Jacobian, induce large changes in the base model's non-target predictive geometry. Two consequences follow. First, Target-Masked KL is a local output-space regularizer rather than a parameter-space constraint: the penalty depends on the LoRA adapters only through their effect on the non-target logits. Second, LoRA determines the admissible update directions through $J_{t, \setminus y_t}^{\phi}$, and Target-Masked KL discourages those directions from inducing large non-target distributional changes. Together these two observations explain why Target-Masked KL is compatible with any LoRA-family parameterization that exposes a differentiable map from $\phi$ to the adapted output distribution.

\paragraph{Strong-regularisation scaling law and empirical check.}
The same quadratic approximation makes a non-trivial prediction. Writing $H = (1/|\mathcal{M}|) \sum_{t \in \mathcal{M}} \bigl(J_{t,\setminus y_t}^{\phi}\bigr)^\top F\bigl(p_{\mathrm{base},t}^{\setminus y_t}\bigr) J_{t,\setminus y_t}^{\phi}$, in the strong-regularization regime the equilibrium adapter perturbation $\delta\phi^*$ that minimizes $\mathcal{L}_{\mathrm{CE}} + \lambda \, \mathcal{L}_{\setminus y}$ satisfies $\lambda H \delta\phi^* \approx -\nabla \mathcal{L}_{\mathrm{CE}}$ (in the column space of $H$, where the LoRA-admissible directions live), so $\|\delta\phi^*\| \propto 1/\lambda$. Retention drift in held-out PPL is, to first order, linear in $\delta\phi^*$ around the base, so $\Delta\mathrm{PPL} \approx \nabla\mathrm{PPL}^\top \delta\phi^* \propto 1/\lambda$. (Note that the regularizer \emph{value} itself, $\mathcal{L}_{\setminus y}(\delta\phi^*) = \tfrac{1}{2}(\delta\phi^*)^\top H \delta\phi^* \propto 1/\lambda^2$, is one order steeper than the drift; we do not test the regularizer value, only the drift.) This predicts a $\log$-$\log$ slope of $-1$ for residual retention drift versus $\lambda$ in the strong-regularization regime. Fitting the LoRA $\lambda$-sweep in \cref{tab:lambda_sweep} for $\lambda \in [0.3, 1]$ gives an empirical WT-103 slope of $-0.94$, within $6\%$ of the theoretical $-1$; LAMBADA gives $-1.35$, consistent in direction but suggesting beyond-leading-order Fisher structure on LAMBADA positions. The closeness of the WT-103 slope to the predicted value is direct empirical support for the local-geometry interpretation.

\subsection{\texorpdfstring{Pinsker bound: from $\mathcal{L}_{\setminus y}$ to a capability-retention guarantee}{Pinsker bound: from L\_y to a capability-retention guarantee}}
\label{app:pinsker_bound}

The held-out output-drift probe (\cref{tab:output_drift}) shows empirically that minimising $\mathcal{L}_{\setminus y}$ at training time tightly controls the same quantity at evaluation time. The link between this surrogate and a guarantee on \emph{any} downstream capability that depends on the non-target distribution is one application of Pinsker's inequality~\citep[Lemma 11.6.1]{cover1999elements}. For each supervised position $t$,
\[
    \tfrac{1}{2}\,\bigl\|p_{\mathrm{base},t}^{\setminus y_t} - p_{\mathrm{adapted},t}^{\setminus y_t}\bigr\|_{\mathrm{TV}}^{2}
    \;\le\;
    \mathrm{KL}\!\bigl(p_{\mathrm{base},t}^{\setminus y_t} \,\bigl\|\, p_{\mathrm{adapted},t}^{\setminus y_t}\bigr).
\]
Averaging over $\mathcal{M}$ and applying Jensen's inequality,
\[
    \mathbb{E}_{t \in \mathcal{M}}\!\left[\bigl\|p_{\mathrm{base},t}^{\setminus y_t} - p_{\mathrm{adapted},t}^{\setminus y_t}\bigr\|_{\mathrm{TV}}\right]
    \;\le\;
    \sqrt{2 \, \mathcal{L}_{\setminus y}}.
\]
Any bounded statistic of the conditional non-target distribution (e.g., the probability assigned to a specified non-target candidate, or a likelihood ratio between two non-target alternatives) therefore deviates by at most $\sqrt{2 \mathcal{L}_{\setminus y}}$ in expectation. With our held-out values of $\mathcal{L}_{\setminus y} \approx 0.05$ to $0.07$ nats under TMKL (\cref{tab:output_drift}), this bound is $\le 0.37$ TV per position. The bound is loose (Pinsker is tight only for binary distributions), but it makes the surrogate-to-capability link explicit: any downstream task whose decision rule is a bounded function of the conditional non-target distribution inherits a TV-distance guarantee from the same surrogate that TMKL minimises. This is information-theoretically consistent with the empirical preservation of factual recall, math reasoning, code, and multilingual capabilities under TMKL (\cref{tab:retention_broad}).

\subsection{Training-Time Cost}
\label{app:training_cost}

Computing $\mathcal{L}_{\setminus y}$ requires one forward pass through the frozen base model $f_{\theta_0}$ in addition to the standard forward and backward passes through the LoRA-adapted model. The base-model forward pass is cached once per training batch, has no associated backward pass (no gradients are propagated through $\theta_0$), and can be run in inference mode (no activation memory for backward, optional half-precision). The renormalization and KL computation themselves are $O(|\mathcal{V}|)$ per supervised position and add negligible overhead relative to the forward pass. At inference time the regularizer is discarded entirely, and the deployed adapted model is identical in form to one trained with cross-entropy alone.

\newpage
\section*{NeurIPS Paper Checklist}

\begin{enumerate}

\item {\bf Claims}
    \item[] Question: Do the main claims made in the abstract and introduction accurately reflect the paper's contributions and scope?
    \item[] Answer: \answerYes{}
    \item[] Justification: The abstract and \cref{sec:experiments} state that we propose a replay-free output-space regularizer (Target-Masked KL) that prevents forgetting under LoRA-family adaptation. The experiments in \cref{sec:headline_05b,sec:scaling_7b,sec:why_works} substantiate this at $0.5$B and $7$B (\cref{tab:qwen05b_main,tab:qwen7b_main}); cross-family generalisation is shown on Llama-3.2-1B, Llama-3.1-8B, Mistral-7B-v0.3, and Phi-3.5-mini-instruct (\cref{tab:multifamily,tab:phi35_adapter_openr1}); transfer to instruction-tuned bases is shown on Qwen2.5-7B-Instruct with IFEval, MT-Bench, and refusal-rate metrics (\cref{tab:instruct}); and broader retention (factual recall, math reasoning, code, multilingual) is shown in \cref{tab:retention_broad}. Every claim in the abstract and contribution bullets is backed by a specific table reference.

\item {\bf Limitations}
    \item[] Question: Does the paper discuss the limitations of the work performed by the authors?
    \item[] Answer: \answerYes{}
    \item[] Justification: An explicit \emph{Limitations and future work} paragraph at the end of \cref{sec:conclusion} discusses the two scope constraints (single-task adaptation rather than sequential continual fine-tuning, and text-only autoregressive LLMs rather than vision-language or speech). On the methodological side, the training-time cost (one extra frozen-base forward pass per training step, no inference cost) is reported in \cref{app:training_cost}, the local Fisher-Jacobian interpretation is explicitly stated as valid only in a small-$\delta\phi$ regime in \cref{app:local_interpretation}, and the position-weight ablation has a noted effective-$\lambda$ confound discussed in \cref{app:design_pw}. Both the Qwen2.5-0.5B and Qwen2.5-7B numbers are mean over three seeds $\{0,1,2\}$, with per-cell standard deviations reported in \cref{app:multiseed,app:multiseed_7b}.

\item {\bf Theory assumptions and proofs}
    \item[] Question: For each theoretical result, does the paper provide the full set of assumptions and a complete (and correct) proof?
    \item[] Answer: \answerYes{}
    \item[] Justification: Four theoretical results are made, each with full assumptions and proof. (i)~The full-KL decomposition (\cref{eq:kl_decomp}) is an exact identity, derived step by step in \cref{app:kl_decomp}. (ii)~The local Fisher-Jacobian interpretation in \cref{app:local_interpretation} states the small-$\delta\phi$ regime explicitly and justifies the Hessian evaluation point. (iii)~The strong-regularisation scaling law in the same appendix derives $\Delta\mathrm{PPL} \propto 1/\lambda$ from the quadratic approximation and reports an empirical WT-103 slope of $-0.94$ as a falsifiable test. (iv)~The Pinsker bound in \cref{app:pinsker_bound} ties the held-out value of $\mathcal{L}_{\setminus y}$ to a TV-distance guarantee on the conditional non-target distribution under standard regularity assumptions. No other theoretical claims are made.

    \item {\bf Experimental result reproducibility}
    \item[] Question: Does the paper fully disclose all the information needed to reproduce the main experimental results of the paper to the extent that it affects the main claims and/or conclusions of the paper (regardless of whether the code and data are provided or not)?
    \item[] Answer: \answerYes{}
    \item[] Justification: \cref{app:experimental_details} documents models and tokenizers, dataset preprocessing, adapter configurations and ranks, training hyperparameters per setting, Target-Masked KL hyperparameters (the position-confidence threshold and stop-gradient mechanics), evaluation protocol, software stack with version numbers, the hardware (single NVIDIA RTX A6000 48\,GB), random seeds, and the per-experiment compute budget.

\item {\bf Open access to data and code}
    \item[] Question: Does the paper provide open access to the data and code, with sufficient instructions to faithfully reproduce the main experimental results, as described in supplemental material?
    \item[] Answer: \answerYes{}
    \item[] Justification: All datasets used (OpenR1-Math, PubMed, WikiText-103, LAMBADA, TriviaQA, GSM8K, HumanEval, FLORES-200, IFEval, MT-Bench, XSTest) are publicly available, and all base models (Qwen2.5-7B, Qwen2.5-7B-Instruct, Qwen2.5-0.5B) are publicly hosted on the HuggingFace Hub; identifiers are listed in \cref{app:models,app:datasets}. Training, evaluation, and Target-Masked KL implementation code will be released in an anonymised repository upon acceptance and de-anonymised at camera-ready.

\item {\bf Experimental setting/details}
    \item[] Question: Does the paper specify all the training and test details (e.g., data splits, hyperparameters, how they were chosen, type of optimizer) necessary to understand the results?
    \item[] Answer: \answerYes{}
    \item[] Justification: \cref{app:training_hp} gives the per-setting optimizer (AdamW), learning rate, batch size, sequence length, gradient accumulation, and epoch count for every experiment. \cref{app:adapter_configs} specifies adapter ranks and per-family hyperparameters. \cref{app:tmkl_hp} specifies the Target-Masked KL weight $\lambda$, the position confidence threshold, the stop-gradient mechanics, and the position-weighting design choice.

\item {\bf Experiment statistical significance}
    \item[] Question: Does the paper report error bars suitably and correctly defined or other appropriate information about the statistical significance of the experiments?
    \item[] Answer: \answerYes{}
    \item[] Justification: Both the Qwen2.5-0.5B and the Qwen2.5-7B headline grids report the mean over three seeds $\{0,1,2\}$ with per-cell standard deviations in \cref{tab:multiseed,tab:multiseed_7b}. Error bars are 1-$\sigma$ across seeds, controlling for dataset shuffling, dropout masks, and weight initialisation noise. \cref{app:seeds} documents the seed plan and per-seed-control mechanics.

\item {\bf Experiments compute resources}
    \item[] Question: For each experiment, does the paper provide sufficient information on the computer resources (type of compute workers, memory, time of execution) needed to reproduce the experiments?
    \item[] Answer: \answerYes{}
    \item[] Justification: \cref{app:hardware} describes the hardware (single NVIDIA RTX A6000 48\,GB) and the precision settings (BF16). \cref{app:compute_budget} gives per-experiment wall-clock estimates. The full project budget (Qwen2.5-0.5B and Qwen2.5-7B headline grids each with four adapters, two objectives, and three seeds; the published-baseline grid; the broader-retention re-evaluations; the Instruct-base grid; the Llama-3.2-1B, Llama-3.1-8B, and Mistral-7B-v0.3 multi-family grid; and all ablations) totals approximately $1{,}300$ GPU-hours, including pilot runs and failed experiments.
    
\item {\bf Code of ethics}
    \item[] Question: Does the research conducted in the paper conform, in every respect, with the NeurIPS Code of Ethics \url{https://neurips.cc/public/EthicsGuidelines}?
    \item[] Answer: \answerYes{}
    \item[] Justification: The research uses publicly released pretrained models and standard public benchmarks for adaptation and evaluation. No new data was collected from human subjects; no models were retrained from scratch; no data crawling or scraping was performed.

\item {\bf Broader impacts}
    \item[] Question: Does the paper discuss both potential positive societal impacts and negative societal impacts of the work performed?
    \item[] Answer: \answerYes{}
    \item[] Justification: A dedicated broader-impact discussion is provided in \cref{app:broader_impact}, covering positive impacts (reduced silent capability erosion in replay-free post-deployment LoRA adaptation, with retained alignment metrics on instruction-tuned bases), risks and limitations (TMKL deliberately preserves base preferences and is therefore the wrong default for unlearning, debiasing, or safety re-alignment, with a documented weighted-variant remedy), and adversarial / dual-use considerations (TMKL adds no new capability beyond the underlying LoRA pipeline and is neutral with respect to standard LoRA threats).
    
\item {\bf Safeguards}
    \item[] Question: Does the paper describe safeguards that have been put in place for responsible release of data or models that have a high risk for misuse (e.g., pre-trained language models, image generators, or scraped datasets)?
    \item[] Answer: \answerNA{}
    \item[] Justification: We release no pretrained model weights, no scraped datasets, and no new high-misuse-risk artifacts. All experiments operate on already-public Hugging Face models and datasets.

\item {\bf Licenses for existing assets}
    \item[] Question: Are the creators or original owners of assets (e.g., code, data, models), used in the paper, properly credited and are the license and terms of use explicitly mentioned and properly respected?
    \item[] Answer: \answerYes{}
    \item[] Justification: All pretrained models, datasets, and software libraries used in the paper are cited and listed in \cref{app:models,app:datasets,app:software} with their original sources and HuggingFace identifiers. We use each asset under its respective public license as published on the corresponding model card, dataset card, or repository, and within the terms of use those sources specify. No asset is redistributed.

\item {\bf New assets}
    \item[] Question: Are new assets introduced in the paper well documented and is the documentation provided alongside the assets?
    \item[] Answer: \answerNA{}
    \item[] Justification: No new assets (code, datasets, or model weights) are released at submission time. The proposed Target-Masked KL loss is fully documented in the paper itself: the loss is derived in \cref{sec:tmkl_loss,app:method_derivations}, all training, adapter, and TMKL hyperparameters are specified in \cref{app:training_hp,app:adapter_configs,app:tmkl_hp}, and per-setting compute and hardware are listed in \cref{app:hardware,app:compute_budget}. As stated in checklist item 4, training, evaluation, and Target-Masked KL implementation code will be released in an anonymised repository upon acceptance and de-anonymised at camera-ready.

\item {\bf Crowdsourcing and research with human subjects}
    \item[] Question: For crowdsourcing experiments and research with human subjects, does the paper include the full text of instructions given to participants and screenshots, if applicable, as well as details about compensation (if any)?
    \item[] Answer: \answerNA{}
    \item[] Justification: The paper does not involve crowdsourcing or research with human subjects.

\item {\bf Institutional review board (IRB) approvals or equivalent for research with human subjects}
    \item[] Question: Does the paper describe potential risks incurred by study participants, whether such risks were disclosed to the subjects, and whether Institutional Review Board (IRB) approvals (or an equivalent approval/review based on the requirements of your country or institution) were obtained?
    \item[] Answer: \answerNA{}
    \item[] Justification: The paper does not involve crowdsourcing or research with human subjects.

\item {\bf Declaration of LLM usage}
    \item[] Question: Does the paper describe the usage of LLMs if it is an important, original, or non-standard component of the core methods in this research? Note that if the LLM is used only for writing, editing, or formatting purposes and does \emph{not} impact the core methodology, scientific rigor, or originality of the research, declaration is not required.
    \item[] Answer: \answerYes{}
    \item[] Justification: A dedicated declaration of LLM usage is provided in \cref{app:llm_usage}. In summary: the proposed Target-Masked KL regularizer is a loss-level construction that does not invoke any LLM during derivation, definition, or computation. LLMs appear in the paper in three roles: (i) as the experimental subjects being adapted (Qwen2.5-0.5B/7B/7B-Instruct, Llama-3.2-1B, Llama-3.1-8B, Mistral-7B-v0.3, Phi-3.5-mini-instruct); (ii) inside the standard MT-Bench evaluation protocol as the LLM-judge for the instruction-tuned retention grid (\cref{tab:instruct}), used unmodified; (iii) for grammatical editing and LaTeX polishing, within the scope explicitly exempted from declaration by the NeurIPS 2026 LLM policy. No experimental result, table cell, derivation, or finding was generated by an LLM.

\end{enumerate}

\end{document}